\documentclass[10pt,twocolumn,letterpaper]{article}

 \usepackage{cvpr}              
\usepackage[accsupp]{axessibility}
\usepackage{threeparttable}

\definecolor{cvprblue}{rgb}{0.21,0.49,0.74}
\usepackage[pagebackref,breaklinks,colorlinks,allcolors=cvprblue]{hyperref}

\def\paperID{WDFM-AD-7} 
\def\confName{CVPR}
\def\confYear{2025}

\title{CleanMAP: Distilling Multimodal LLMs for Confidence-Driven Crowdsourced HD Map Updates}

\author{
Ankit Kumar Shaw\textsuperscript{1} \quad
Kun Jiang\textsuperscript{*1} \quad
Tuopu Wen\textsuperscript{1} \quad
Chandan Kumar Sah\textsuperscript{2} \quad
Yining Shi\textsuperscript{1} \\
Mengmeng Yang\textsuperscript{*1} \quad
Diange Yang\textsuperscript{*1} \quad
Xiaoli Lian\textsuperscript{2} \\
\textsuperscript{1}Tsinghua University \quad
\textsuperscript{2}Beihang University \\
\textsuperscript{*}Corresponding author \\
\tt\small \{shawak10, syn21\}@mails.tsinghua.edu.cn, \{jiangkun, yangmm\_qh\}@tsinghua.edu.cn, \\
\tt\small ydg@mail.tsinghua.edu.cn, wtp18@tsinghua.org.cn, \{sahchandan98, lianxiaoli\}@buaa.edu.cn
}

\begin{document}
\maketitle

\begin{abstract}
The rapid growth of intelligent connected vehicles (ICVs) and integrated vehicle-road-cloud systems has increased the demand for accurate, real-time HD map updates. However, ensuring map reliability remains challenging due to inconsistencies in crowdsourced data, which suffer from motion blur, lighting variations, adverse weather, and lane marking degradation. This paper introduces CleanMAP, a Multimodal Large Language Model (MLLM)-based distillation framework designed to filter and refine crowdsourced data for high-confidence HD map updates. CleanMAP leverages an MLLM-driven lane visibility scoring model that systematically quantifies key visual parameters, assigning confidence scores (0–10) based on their impact on lane detection. A novel dynamic piecewise confidence-scoring function adapts scores based on lane visibility, ensuring strong alignment with human evaluations while effectively filtering unreliable data. To further optimize map accuracy, a confidence-driven local map fusion strategy ranks and selects the top-$k$ highest-scoring local maps within an optimal confidence range (best score minus 10\%), striking a balance between data quality and quantity. Experimental evaluations on a real-world autonomous vehicle dataset validate CleanMAP’s effectiveness, demonstrating that fusing the top three local maps achieves the lowest mean map update error of 0.28m, outperforming the baseline (0.37m) and meeting stringent accuracy thresholds ($\leq$0.32m). Further validation with real-vehicle data confirms 84.88\% alignment with human evaluators, reinforcing the model’s robustness and reliability. This work establishes CleanMAP as a scalable and deployable solution for Crowdsourced HD Map Updates, ensuring more precise and reliable autonomous navigation. The code will be available at \url{https://Ankit-Zefan.github.io/CleanMap/}.

\end{abstract}

\section{Introduction}
\label{sec:intro}

High-definition (HD) maps are vital for autonomous vehicles, supporting precise localization, path planning, and environmental perception~\cite{liu2019hd, seif2016hd}. Unlike traditional maps, HD maps offer centimeter-level accuracy by capturing detailed road elements such as lane boundaries, traffic signs, and static infrastructure, which ensures safe navigation even when onboard sensors fail~\cite{lawton2015hd, pan2021hd, xu2021cs, chang2024hd}. 

Frequent map updates are essential to reflect dynamic real-world changes~\cite{guo2024crowdsourcing, kim2023v2i}. However, conventional update methods using dedicated mapping fleets are prohibitively expensive and infeasible at scale~\cite{hortelano2023drivable, mohamed2023lowcost}. Crowdsourced data from connected and automated vehicles (CAVs) offers a scalable alternative for real-time updates~\cite{guo2024review, xu2024towards}. Yet, this data often suffers from quality issues—including motion blur, poor lighting, adverse weather, and sensor noise—that degrade lane visibility and introduce mapping errors~\cite{zhang2023blind, guo2024review}. Since lane markings are critical for localization and path planning~\cite{xu2024towards, guo2024review}, unreliable data can lead to incorrect lane inference and navigation failures.

\begin{figure*}[t]
\centering
    \includegraphics[width=2\columnwidth]{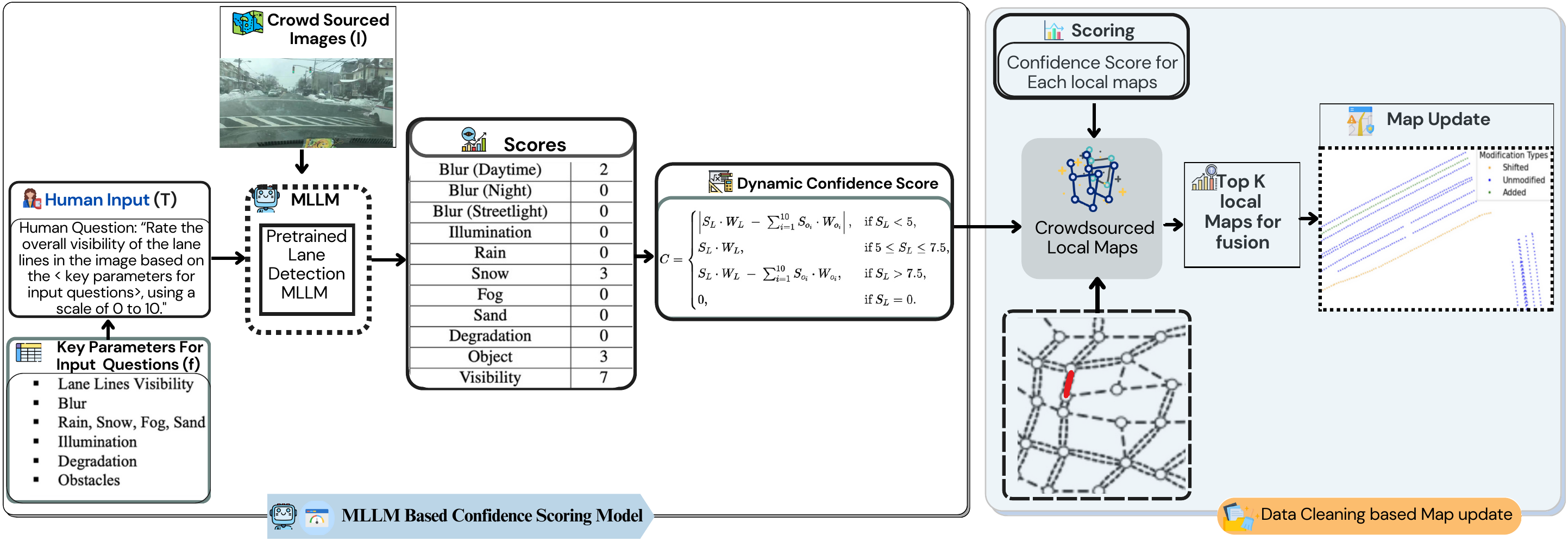}
\caption{Overall Framework for MLLM-driven Data Cleansing for Crowdsourced HD Map Updates. A pretrained MLLM processes multimodal inputs, scoring key parameters related to lane visibility. The confidence-driven selection filters high-quality local maps, ranks them, and fuses the top \( k \) maps within an optimal confidence range to enhance HD map updates.}
\label{fig:A1}
\end{figure*}

Human validation remains the standard for assessing data quality, but it is costly, time-consuming, and unscalable for large-scale deployments~\cite{parameswaran2012}. While CNN-based perceptual quality models evaluate factors like noise and blur~\cite{talebi2018nima}, they fail to address lane-specific visibility. Vision-Language Models (VLMs) improve general image assessment~\cite{chen2023iqagpt, qalign}, yet they focus on holistic clarity rather than visibility of lane features under challenging conditions~\cite{fourati2024xlm}. Furthermore, categorical quality labels (e.g., “good,” “fair,” “poor”)~\cite{qbench, qalign} lack the granularity needed for fine-grained confidence estimation. A robust, quantitative scoring mechanism is needed to ensure high-confidence images contribute to HD map updates, while low-confidence inputs are filtered out~\cite{xu2024towards}. Automating this process improves consistency, scalability, and responsiveness~\cite{fourati2024xlm}.

A further challenge lies in balancing data quality and coverage. While CAVs generate a constant stream of crowdsourced data, inconsistent quality necessitates selective filtering. Overly aggressive filtering risks coverage gaps, while lenient policies introduce noise. Fixed-threshold methods such as zero-shot filtering—may retain poor data or discard valuable samples. In contrast, unfiltered aggregation inflates processing costs and reduces map precision~\cite{pan2021hd, xu2021cs}. These challenges call for an adaptive, confidence-aware data selection strategy.

To address these challenges, we propose CleanMAP, a novel Multimodal Large Language Model (MLLM)-based framework for confidence-driven HD map updates. CleanMAP introduces a fine-grained, MLLM-driven lane visibility scoring model that quantitatively assesses critical visual degradation factors affecting HD map quality. To improve decision accuracy, we design an adaptive, dynamic confidence-scoring mechanism that closely aligns with human perception and robustly filters unreliable data. Further, we propose a confidence-driven local map fusion strategy that adaptively selects the top-$k$ highest-scoring local maps within a relative confidence band (defined as the best score minus 10\%), maintaining an optimal balance between data quality and coverage. Our extensive real-world evaluations demonstrate that CleanMAP significantly improves HD map reliability and precision, offering a scalable and automated solution for continuous updates in autonomous navigation systems.

The key contributions of this work are:

\begin{itemize}
    \item \textbf{MLLM-driven lane visibility scoring:} We develop a structured scoring model that evaluates key visibility degradation factors—including motion blur, illumination, weather conditions, occlusion, and lane wear—assigning each a quantitative score (0–10). This enables precise, lane-specific quality assessment to inform HD map updates.

    \item \textbf{Adaptive confidence-scoring mechanism:} We introduce a novel piecewise dynamic function that translates visibility parameters into confidence scores, enhancing alignment with human judgment and ensuring robust filtering under diverse environmental conditions.

    \item \textbf{Optimal confidence-driven local map fusion:} We present a selective fusion strategy that ranks local maps by confidence score and selects the top-$k$ within an optimal confidence margin (best score - 10\%), effectively balancing data quality and quantity to maximize map accuracy.
\end{itemize}

\section{Related Works}
\label{sec:related}
Prior studies on crowdsourced HD map updates have predominantly focused on data quantity, often neglecting quality assessment, leading to high map update errors. For instance, \cite{guo2024review} highlights the challenge of cleansing and enhancing crowdsourced data quality, emphasizing the need for robust filtering mechanisms. Existing crowdsourcing approaches \cite{liebner2019crowdsourced, das2020posegraph, kim2021hdmapupdate} have primarily relied on direct updates focusing on quantity only, leading to map update inaccuracies. This tradeoff between quality and quantity underscores the need for quality-driven methods, which can optimize both dimensions.

\subsection{Human-Based Data Quality Assessment}
Human-in-the-loop (HITL) approaches remain critical for HD map verification, particularly for lane markings, road boundaries, and traffic signs. Hybrid validation models, integrating human expertise with automation, have been explored in \cite{guo2024review}. The HITL paradigm \cite{wu2022survey} as well as human-based quality assessment methods, such as CrowdScreen \cite{parameswaran2012crowdscreen}, rely on manual annotations to validate data quality and enhance reliability but suffer from slow execution, human bias, and high costs. Multi-sensor fusion techniques \cite{hossain2011modeling} propose confidence-based quality assessments, reinforcing the necessity of human validation in refining HD maps.

\subsection{Deep Learning-Based Data Quality Assessment}
Deep learning has significantly advanced automated quality assessments, particularly in image perceptual quality estimation. Early works in Image Quality Assessment (IQA), such as \cite{wang2004image, mittal2012niqe, mittal2013blique}, introduced handcrafted feature-based approaches. Recent neural network-driven models like NIMA \cite{talebi2018nima}, DBCNN \cite{zhang2020dbcnn}, and HyperIQA \cite{su2020hyperiqa} have improved generalization capabilities. The MUSIQ framework \cite{ke2021musiq} further refines image quality prediction using transformer architectures. Studies on blind IQA \cite{zhao2023quality, gu2022ntire} leverage self-supervised contrastive learning, while research in perceptual IQA \cite{zhang2022perceptual} highlights robustness challenges in deep-learning-based models. Despite these advances, traditional IQA primarily assesses noise, blur, and compression artifacts, which are insufficient for lane visibility prioritization in HD maps.

\subsection{VLM-Based Visual Scoring}
Recent breakthroughs in VLMs have enabled multimodal quality assessment by integrating textual and visual reasoning. CLIP-based IQA models \cite{radford2021clip, wang2022clipiqa, zhang2023liqe} exploit vision-language correspondence for general image assessment. The IQAGPT framework \cite{chen2023iqagpt} introduces VLM-based textual justifications alongside image quality scoring. Blind image quality assessment via vision-language fusion \cite{zhang2023blind} classifies distortions and scene attributes, complementing CleanMAP’s multimodal approach. Further studies, such as \cite{li2023seed, zhu20242afc, xu2024towards, qbench, qalign}, emphasize structured multimodal quality assessment for improved decision-making. However, conventional VLM-based scoring lacks fine-grained confidence estimation, often categorizing images as "good," "fair," or "poor," which is insufficient for HD map filtering under adverse conditions.

Despite these advancements, existing methods fail to integrate task-specific multimodal assessment for HD maps. CleanMAP bridges this gap by leveraging MLLM-driven confidence scoring, ensuring robust lane visibility prioritization and automated confidence ranking for HD map updates. Inspired by Q-Align \cite{qalign}, which enhances structured multimodal quality assessment, CleanMAP introduces a confidence-driven filtering mechanism, balancing quality and quantity in crowdsourced map updates.

\section{Methodology}
\label{sec:method}
This work presents a Multimodal LLM-Driven Confidence Scoring Model for automated quality assessment of crowdsourced HD map data, ensuring that only high-confidence images contribute to map updates. As illustrated in Figure \ref{fig:A1}, the proposed framework utilizes a pretrained Multimodal Large Language Model (MLLM) from \cite{sah2025auto}, designed to extract key visual information related to lane line visibility under adverse environmental conditions. The model evaluates images based on key parameters such as motion blur, adverse weather, degradation, and poor illumination, distilling this knowledge into quantitative scores. These scores are further processed using a dynamic piecewise confidence-scoring mechanism to compute a final confidence score for each image. Leveraging these scores, the framework selects optimal local maps for fusion in the HD map update system, ensuring a balance between data quality and quantity while enhancing the accuracy and reliability of HD maps.


\begin{figure}[t]
\centering
    \includegraphics[width=1.0\columnwidth]{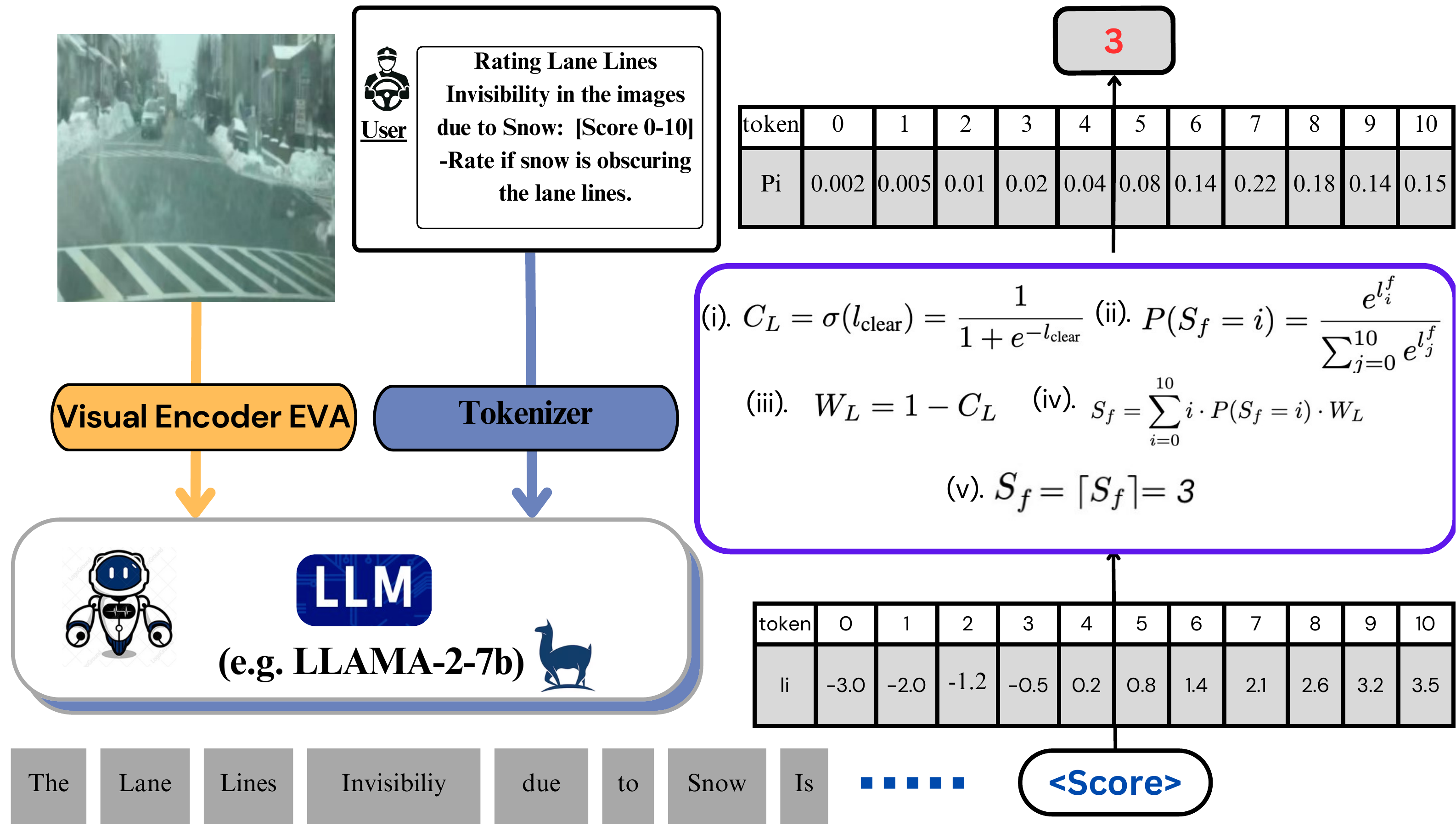}
\caption{MLLM-driven scoring model for evaluating key parameters related to lane line visibility in an image.}
\label{fig:M1}
\end{figure}

It is noted that the parameters irrelevant to the specific condition (e.g. fog during a sunny day) are either assigned low weights or ignored to prevent confusion in the model’s reasoning process.



\subsection{MLLM-driven Scoring}

Inspired by prior works \cite{xu2024towards, qalign, qbench}, we propose an MLLM-driven scoring model for crowdsourced data quality assessment, leveraging a lane detection-pretrained MLLM \cite{sah2025auto}. This pretrained model integrates EVA \cite{eva} as the vision encoder and LLaMA 2-7B \cite{llama2} as the LLM and is further optimized using LoRA \cite{lora}. Given an input image and structured text-based queries, the model predicts a quantitative visibility score (0-10) for each factor affecting lane visibility.

\subsubsection{Vision and Text Encoding}

For an input image \( I \) and textual annotation \( T \), the EVA encoder extracts visual features \( F_V \)(see Figure~\ref{fig:A1}), while LLAMA 2-7B tokenizes and encodes the textual input into embeddings \( F_T \). The extracted representations are formulated as:

\begin{equation}
    F_V = \text{EVA}(I), \quad F_T = \text{LLAMA}(T)
\end{equation}

The joint vision-language representation is computed as:

\begin{equation}
    F = \text{MLLM}(F_V, F_T)
\end{equation}

where \( F_V \) represents the vision feature extracted from EVA, \( F_T \) is the text embedding generated by LLAMA 2-7B, and \( F \) is the fused multi-modal representation.

\subsubsection{Logit Computation for Each Score}

For each visibility factor \( f \), LLAMA 2-7B generates logits \( l_i^f \) for scores \( S_f = i \), where \( i \) ranges from 0 to 10. 
\begin{equation}
    f \in \{ B_D, B_N, B_{NL}, I, R, S, F, S_S, O, D\}
\end{equation}

For each visibility-affecting factor \( f \), the MLLM generates raw logits \( l_i^f \) for scores \( S_f = i \), where \( i \) ranges from 0 to 10. The logits are computed as follows:

\begin{equation}
    l_i^f = \text{MLLM}_{\theta}(F, Q_f, i)
\end{equation}

where \( \text{MLLM}_{\theta} \) represents the parameterized Multimodal Large Language Model, \( F \) is the image input, \( Q_f \) is the structured prompt corresponding to the factor \( f \), \( i \) is the score level from 0 to 10, and \( l_i^f \) is the raw logit output for score \( i \). This formulation allows the model to evaluate each visibility-affecting factor independently.

\subsubsection{Lane Feature Detection Confidence Computation}
Since the MLLM processes both textual and visual inputs, the lane visibility confidence \( C_L \) is extracted using an explicit vision-language prompting approach. Given an image \( F \), the MLLM is prompted with:

\textit{"How clearly are the lane markings visible in this image? Rate from 0 (completely invisible) to 1 (fully visible)."}

The model generates logits \( l_{\text{clear}} \) corresponding to the word "clear," representing the confidence of lane visibility. A sigmoid activation function is applied to normalize this into a probability:

\begin{equation}
    C_L = \sigma(l_{\text{clear}}) = \frac{1}{1 + e^{-l_{\text{clear}}}}
\end{equation}

where \( l_{\text{clear}} \) is the logit score from the MLLM for "clear." The sigmoid function ensures that \( C_L \) remains within the range [0,1], making it a valid confidence measure.

\subsubsection{Discrete Lane Visibility Score Computation}
Once \( C_L \) is computed, it is transformed into a discrete integer score \( S_L \) ranging from 0 to 10:

\begin{equation}
    S_L = \text{round}(10 \cdot C_L)
\end{equation}

This transformation ensures that \( S_L \) is a whole number between 0 and 10. Higher values of \( C_L \) correspond to higher lane visibility scores. The computed score is interpretable and consistent with human annotations.

\subsubsection{Softmax Probability Computation}
To convert logits into a probability distribution, the softmax function is applied:

\begin{equation}
    P(S_f = i) = \frac{e^{l_i^f}}{\sum_{j=0}^{10} e^{l_j^f}}
\end{equation}

where \( P(S_f = i) \) is the probability of assigning score \( i \). The softmax function ensures a valid probability distribution, where all probabilities sum to 1 across the possible scores ranging from 0 to 10. This allows the model to compute a probabilistic estimate for the severity of each visibility factor \( f \) affecting lane visibility.

\subsubsection{Visibility-Aware Confidence Weighting (VACW)}
To ensure that the scoring model prioritizes lane visibility rather than overall image degradation, a lane visibility confidence weight \( W_L \) is introduced. This weight is defined as:

\begin{equation}
    W_L = 1 - C_L
\end{equation}

where \( C_L \) represents the MLLM's lane feature detection confidence, which is a value between 0 and 1. The confidence weight functions as follows. If lane lines are highly visible (\( C_L \approx 1 \)), the weight \( W_L \) is low, reducing the contribution of the visibility factor \( f \) to the final score. If lane lines are poorly visible (\( C_L \approx 0 \)), the weight \( W_L \) is high, increasing the contribution of the factor to the final score. The final visibility-calibrated score for each factor is then computed as:

\begin{equation}
    S_f = \sum_{i=0}^{10} i \cdot P(S_f = i) \cdot W_L
\end{equation}

This ensures that visibility-related scores are weighted appropriately, aligning with the model's detection confidence of lane clarity.

\subsubsection{Integer Normalization for Final Score}
Since absolute integer values between 0 and 10 are required for practical use, an integer normalization step is applied:

\begin{equation}
    S_f^{\text{final}} = \lceil S_f \rceil
\end{equation}

Thus, the final score for each factor always belongs to the set:

\begin{equation}
    S_f^{\text{final}} \in \{0,1,2,\dots,10\}
\end{equation}

This rounding step ensures that the computed scores align with human-level interpretability and remain consistent with manual annotations.

This formulation integrates both visibility factor \( f \)-based scoring and direct lane visibility confidence scoring, ensuring that the final scores are context-aware and reliable for HD map updates.

\subsection{Dynamic Piecewise Confidence Scoring (DPCS)}

To ensure reliable HD map updates, a robust confidence scoring system is essential for evaluating image quality. The confidence score quantifies key parameters to determine whether an image meets the required standard for inclusion in HD maps.

Lane line visibility, represented by \( S_L \), is the primary factor, alongside 10 additional parameters \( S_{O_i} \) (\( i = 1, 2, \dots, 10 \)), each weighted accordingly (\( W_L \) for visibility and \( W_{O_i} \) for other factors). The model dynamically adjusts these weights based on environmental conditions to compute an accurate confidence score.

The final confidence score \( C \) is determined using a dynamic piecewise function, adapting to different visibility levels to ensure optimal image selection for HD map updates.

\begin{equation}
    C =
    \begin{cases} 
        |S_L \cdot W_L - \sum\limits_{i=1}^{10} S_{O_i} \cdot W_{O_i}|, & \text{if } S_L < 5 \\ 
        S_L \cdot W_L, & \text{if } 5 \leq S_L \leq 7 \\ 
        S_L \cdot W_L - \sum\limits_{i=1}^{10} S_{O_i} \cdot W_{O_i}, & \text{if } S_L > 7 \\ 
        0, & \text{if } S_L = 0
    \end{cases}
\end{equation}

For \( S_L < 5 \), low visibility results in a reduced confidence score, as other parameters significantly impact lane clarity.  

For \( 5 \leq S_L \leq 7 \), moderate visibility allows the confidence score to be determined primarily by \( S_L \), while other factors contribute but do not drastically lower the lane lines visibility score, ensuring a balanced assessment.  

For \( S_L > 7 \), high visibility leads to a slight adjustment by subtracting the weighted influence of other parameters. In rare cases, such as heavy rain with clear lane lines, these factors play a role but have a limited effect, maintaining a reliable confidence score.

For \( S_L = 0 \), the confidence score is set to zero, indicating the image is unsuitable for HD map updates due to severe occlusion, extreme blur, or sensor failure.

The human evaluator follows the General Confidence Scoring (GCS) given by:
\begin{equation}
    C =
    \begin{cases} 
         0, & \text{if } S_L = 0 \\
        |S_L \cdot W_L - \sum\limits_{i=1}^{10} S_{O_i} \cdot W_{O_i}|, & \text{otherwise} 
    \end{cases}
\end{equation}

\subsection{Confidence-Optimized Map Fusion Strategy}

Achieving high-accuracy HD map updates requires a confidence-driven fusion strategy that balances data quality and quantity. We distill MLLM-based confidence scores into the update system, selecting the most reliable local maps to minimize errors and enhance robustness.

Each map linklet consists of multiple local maps derived from spatio-temporal image sequences. Confidence scores are computed per image and averaged to rank local maps. To optimize fusion, we select the top-$k$ highest-confidence local maps, ensuring both accuracy and data sufficiency. 

We define the optimal selection range as:

\begin{equation}
    \left[ C_{\text{best}}, C_{\text{best}} - 0.1 C_{\text{best}} \right]
\end{equation}

where $C_{\text{best}}$ is the highest confidence score, with a 10\% reduction setting the lower bound. This strategy refines HD map updates by maximizing precision, minimizing errors, and improving reliability for autonomous navigation.

\section{Experimental Results}
\label{sec:res}

\subsection{Datasets and Annotation}

\begin{figure}[t]
\centering
    \includegraphics[width=0.8\columnwidth]{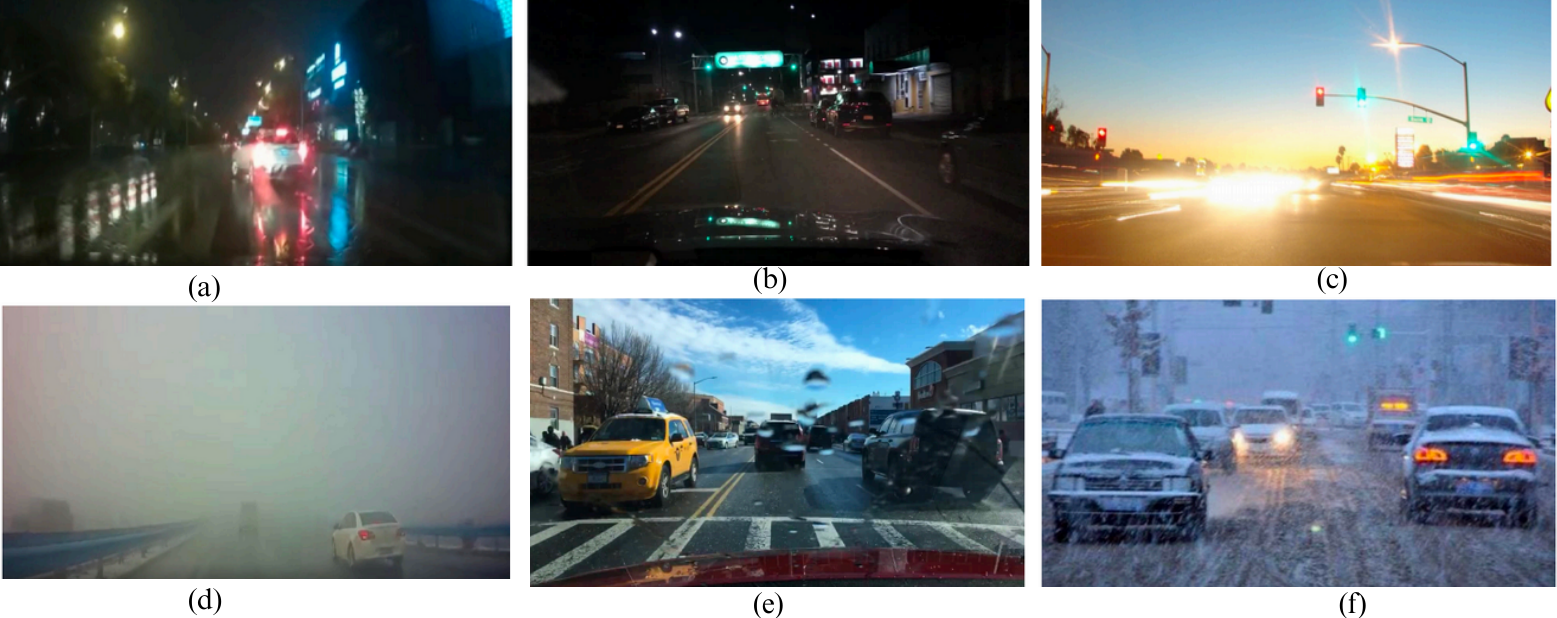}
\caption{Sample training images for instruction-tuning the MLLM-driven scoring model:  
(a) Nighttime blur, (b) Clear nighttime, (c) Glare from streetlights and vehicles, (d) Fog-induced blur, (e) Light rain on a busy street, (f) Heavy snowfall with blur.}
\label{fig:d}
\end{figure}

\begin{figure*}[h]
\centering
    \includegraphics[width=2.0\columnwidth]{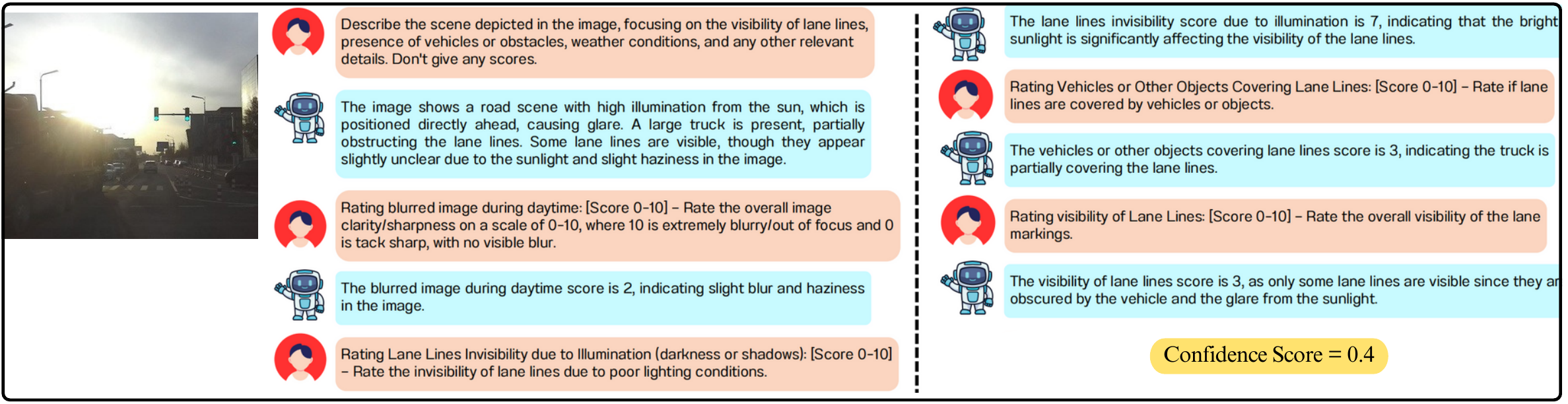}
\caption{MLLM-driven assessment of lane visibility under bright sunlight causing glare and partial occlusion of lane lines by vehicles.}
\label{fig:qa1}
\end{figure*}



The training dataset comprises 10,000 images, integrating online crowdsourced and synthetic data for diverse environmental coverage (Figure \ref{fig:d}), enhancing model generalization for HD map updates.

\begin{figure}[t]
\centering
    \includegraphics[width=0.78 \columnwidth]{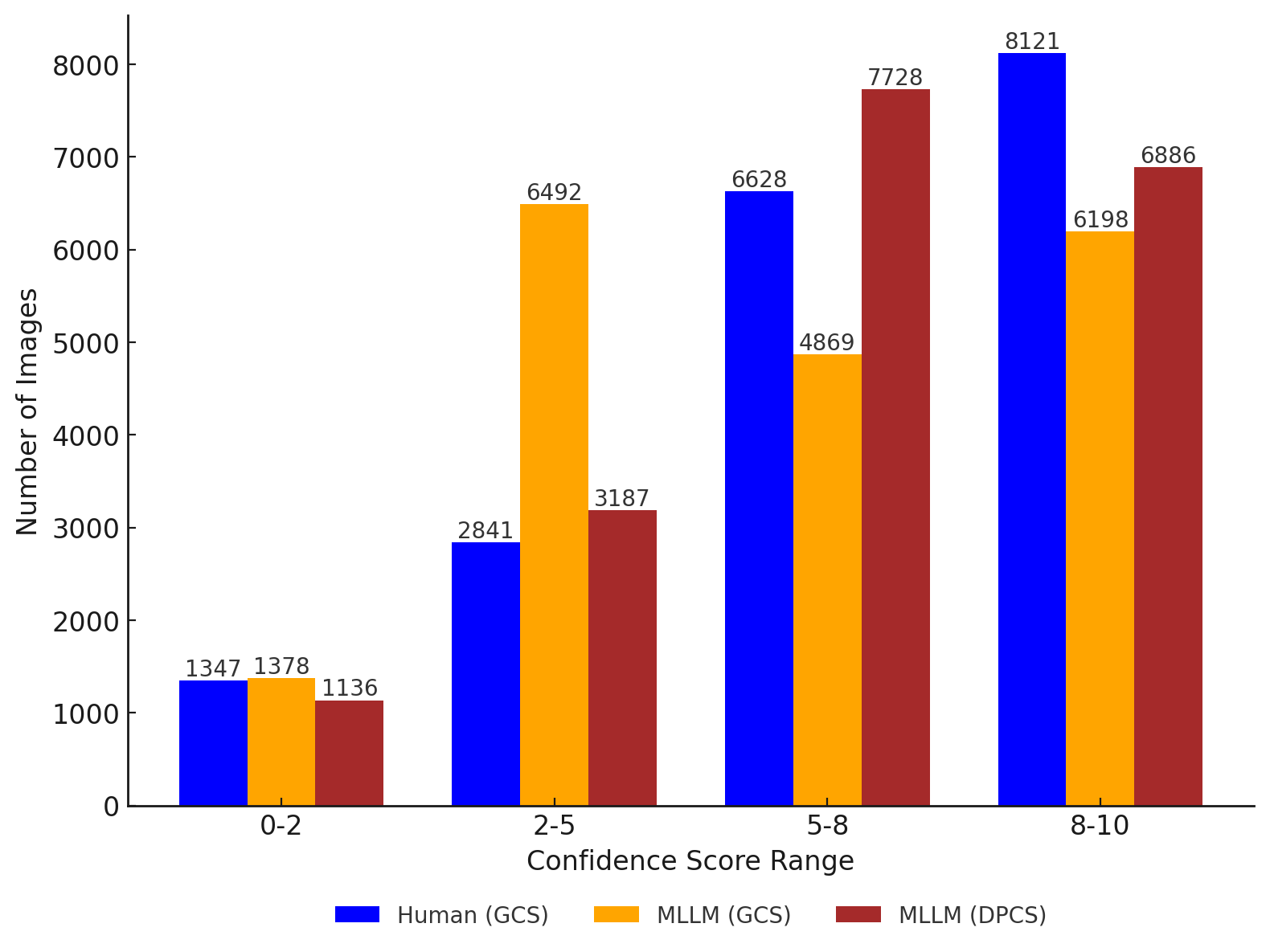}
\caption{Confidence Score Distribution for Human vs Our MLLLM with 18937 Crowdsourced Vehicle Collected timestamp images.}
\label{fig:bar}
\end{figure}

For evaluation, the MLLM-driven confidence scoring model was first tested on 500 randomly selected images from a dataset similar to the training set. Final testing utilized 18,937 real-world images from a Xiaopeng G3 vehicle in Beijing, balancing data richness and computational feasibility.  

Each image was manually annotated for key visual factors like blur, rain, snow, lighting, and occlusions to ensure high-quality assessment. This enabled the model to capture nuanced lane visibility variations, from near-invisible lanes in dense fog to clear visibility in mild rain, ensuring robust real-world performance.

\subsection{Implementation Details}

We employ an MLLM-driven scoring model for crowdsourced data quality assessment, leveraging a lane detection-pretrained MLLM \cite{sah2025auto}. The EVA visual backbone remains frozen, fine-tuning only the linear projection layer for vision-language fusion. LoRA (\( r = 64 \)) is applied to the transformer's query and value projection matrices for efficient adaptation. Training is conducted on 8 NVIDIA RTX 3090 GPUs with 448×448 input images, using AdamW and a cosine learning rate scheduler (\( 1e^{-5} \) max LR, 0.05 warmup). Cross-Entropy Loss optimizes lane detection and instruction-following tasks over 20 epochs, with a batch size of 1 due to computational constraints. A temperature of 0.6 ensures coherent outputs for ambiguous data quality cases.

For confidence score computation, parameter weights are designed to reflect reflect real-world conditions during crowdsourced data collection. The lane line visibility being most important, its weight is set as:
\begin{equation}
    W_L = 1.0
\end{equation}
Other environmental factors, including blur, rain, snow, fog, degradation, and occlusion, share equal weights:
\begin{equation}
    \begin{aligned}
        W_B = W_N = W_{NL} = W_R &= 0.2, \\
        W_S = W_F = W_D = W_O &= 0.2
    \end{aligned}
\end{equation}

Sandstorms, being rare in the dataset, are weighted lower:
\begin{equation}
    W_{SS} = 0.1
\end{equation}

This structured weighting ensures lane visibility remains the dominant factor while allowing adaptive adjustments for extreme conditions.

\subsection{Evaluation Metric for HD Map Updates}

To quantify HD map update accuracy, the root mean square error (RMSE)\cite{xiao2020monocular} is used to compute the Average Mapping Error (AME) relative to the ground truth:
\begin{equation}
    e_{\text{AME}} = \sqrt{\frac{1}{N} \sum_{i=1}^{N} \| X_i - X_i^{\text{GT}} \|^2 }
\end{equation}
where \( i \) represents map points, and \( X_i^{\text{GT}} \) denotes the ground truth. For lane lines, only the lateral coordinates of \( X \) are considered.








\subsection{Results and Discussions}


\begin{table}[h!]
\centering
\caption{Scores Comparison: Our MLLM vs Human Evaluator}
\vspace{-5.50pt}
\label{tab:table_image1}
\resizebox{\columnwidth}{!}
{
\begin{tabular}{@{}llcccccc@{}}
\toprule
\quad\quad & \multicolumn{7}{c}{\textbf{Images}}\\
\cmidrule{3-8}
&& \multicolumn{2}{c}{\includegraphics[width=0.25\linewidth, height=0.2\linewidth]{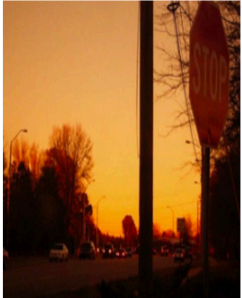}} & \multicolumn{2}{c}{\includegraphics[width=0.25\linewidth, height=0.2\linewidth]{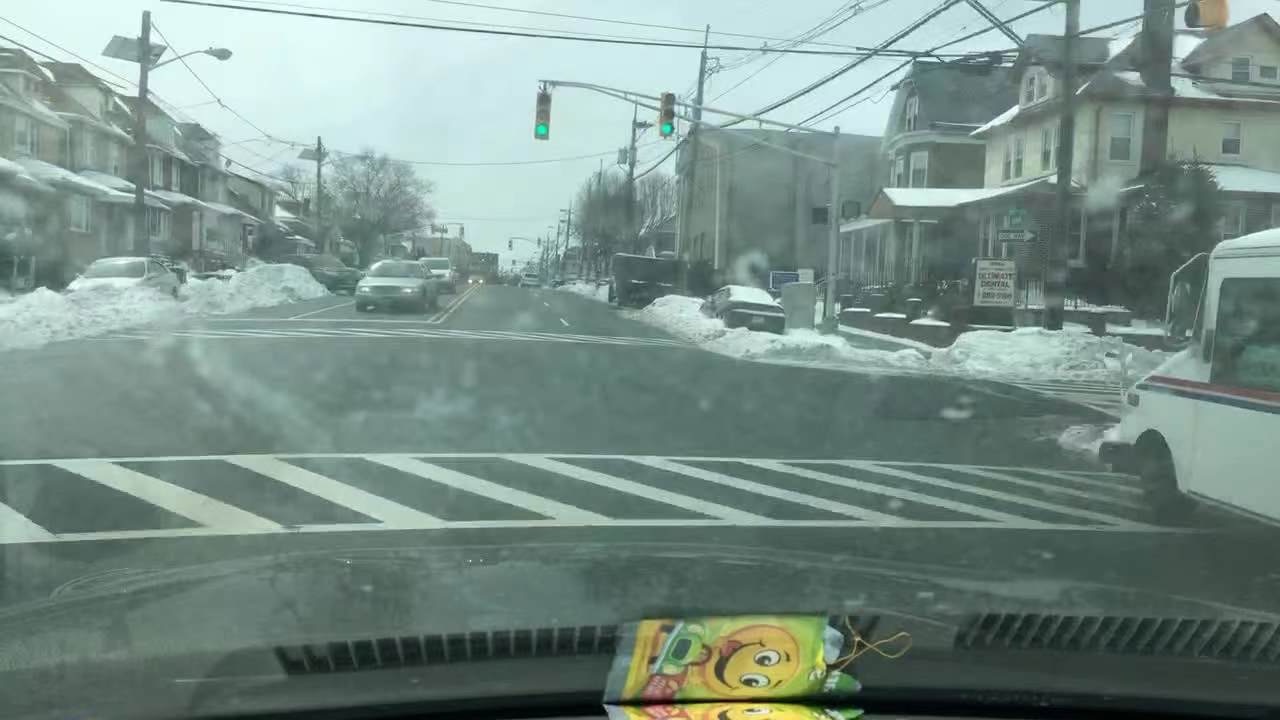}} & \multicolumn{2}{c}{\includegraphics[width=0.25\linewidth, height=0.2\linewidth]{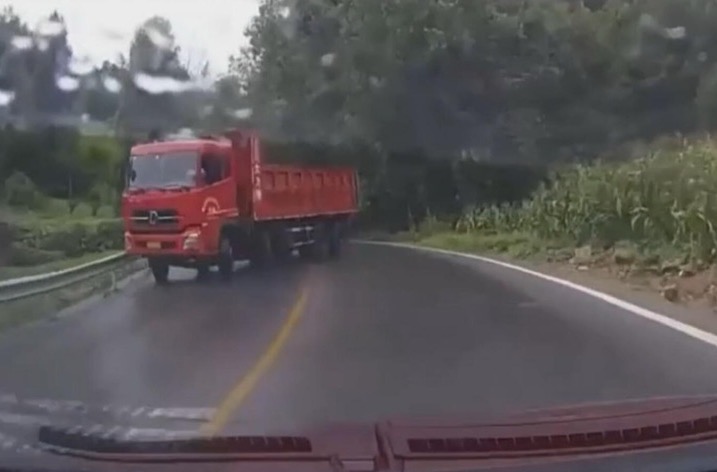}} \\
\cmidrule{3-8}
\textbf{Parameters} && \textbf{Ours} & \textbf{Human} & \textbf{Ours} & \textbf{Human} & \textbf{Ours} & \textbf{Human} \\ 
\midrule
Blur (Daytime)    & & 3 & 4 & 2 & 2 & 2 & 1 \\
Blur (Night)      & & 0 & 0 & 0 & 0 & 0 & 0 \\
Blur (Streetlight)& & 0 & 0 & 0 & 0 & 0 & 0 \\
Illumination      & & 7 & 5 & 0 & 0 & 0 & 0 \\
Rain              & & 0 & 0 & 0 & 0 & 3 & 2 \\
Snow              & & 0 & 0 & 3 & 2 & 0 & 0 \\
Fog               & & 0 & 0 & 0 & 0 & 0 & 0 \\
Sand              & & 2 & 0 & 0 & 0 & 0 & 0 \\
Degradation       & & 0 & 0 & 0 & 0 & 0 & 0 \\
Object Occlusion  & & 2 & 0 & 2 & 1 & 2 & 2 \\
Visibility        & & 0 & 0 & 7 & 8 & 9 & 9 \\
\midrule
{Confidence Score (GCS)} & & 0 & 0 & 5 & 7 & 7.6 & 8 \\
{Confidence Score (DPCS)} & & 0 & - & 7 & - & 7.6 & - \\
\bottomrule
\end{tabular}
}
\begin{tablenotes}
\scriptsize
\item GCS: General Confidence Score and DPCS: Dynamic Piecewise Confidence Score
\end{tablenotes}
\end{table}

\begin{table}[h!]
\centering
\caption{Confidence Score Comparison with different methods}
\vspace{-5.50pt}
\label{tab:mini}
\resizebox{\columnwidth}{!}{
\begin{tabular}{llcccccc}
\toprule
{\quad}\quad & \multicolumn{4}{c}{\textbf{Images}} \\
\cmidrule{3-5}
&& \includegraphics[width=0.3\linewidth, height=0.2\linewidth]{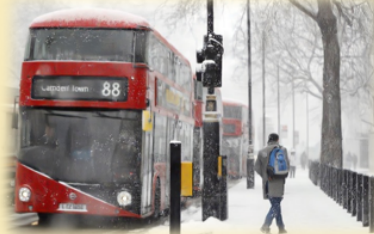} & \includegraphics[width=0.3\linewidth, height=0.2\linewidth]{c.JPEG} & \includegraphics[width=0.3\linewidth, height=0.2\linewidth]{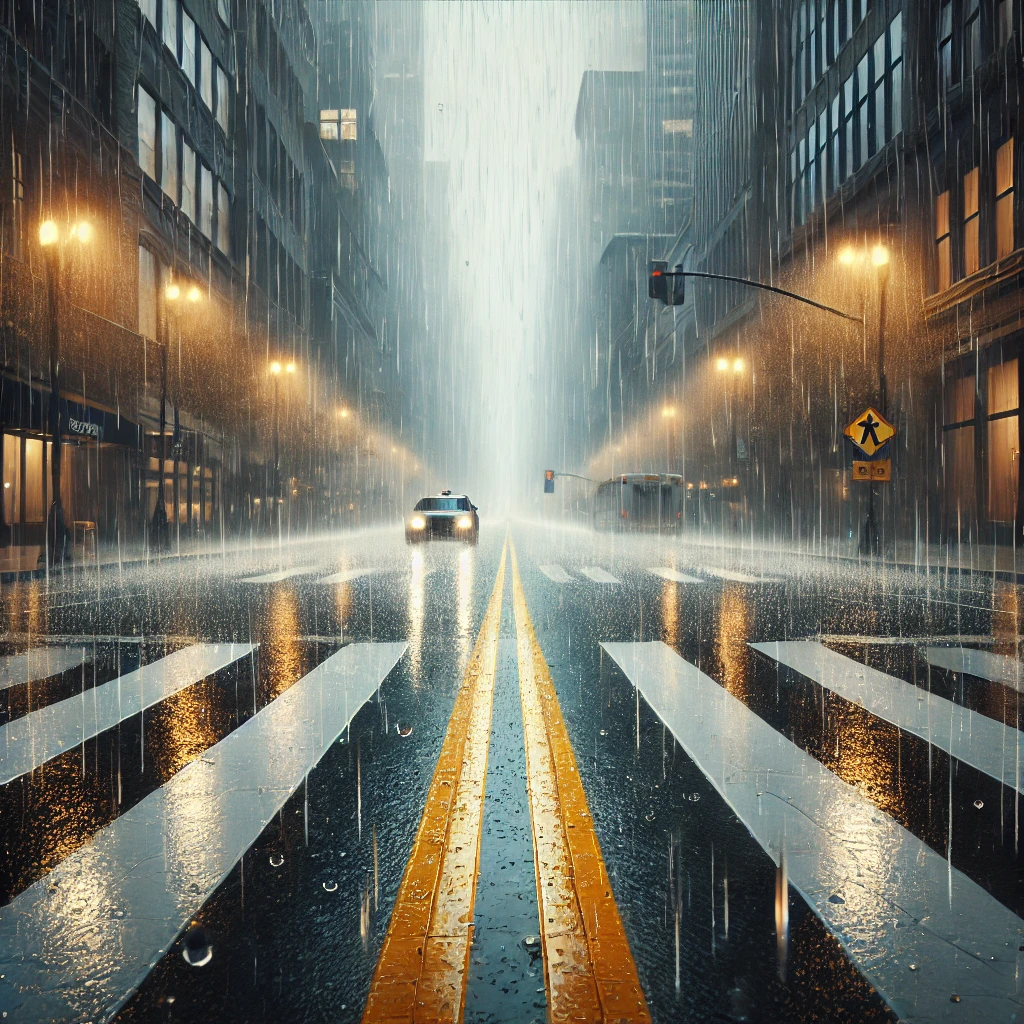} \\
\cmidrule{3-5}
{\textbf{Methods}} & \multicolumn{4}{c}{\textbf{Confidence Score}}\\ 
\midrule
MiniGPT-v2 (zero shot) \cite{minigpt} & & 9 & 5 & 3 \\
GPT-4v (zero shot) \cite{gpt4v} & & 8 & 5 & 4 \\
WResVLM \cite{xu2024towards} & & Good (8) & Fair (5) & Poor (3) \\
Human (GS) & & 0 & 7 & 8.2 \\
Ours MLLM (DPCS) & & 0 & 7 & 7.6 \\
\bottomrule
\end{tabular}
}
\end{table}

\subsubsection{Evaluation on Random Online Crowdsourced Images}


Figure \ref{fig:qa1} demonstrates the effectiveness of the MLLM-driven confidence scoring model in accurately assessing lane visibility under challenging conditions. The model correctly identified the primary obstructions affecting lane clarity, including severe glare from direct sunlight (7/10), partial occlusion by a large truck and other vehicles (3/10), and minor blur (2/10). These factors collectively resulted in a low lane visibility score of 3/10, reinforcing the image’s unsuitability for HD map updates. The confidence score of 0.4 confirms the model's ability to filter out poor-quality images, preventing unreliable data from compromising HD map accuracy. This validates the robustness of the confidence scoring framework in ensuring only high-quality data contributes to precise mapping.

Table \ref{tab:table_image1} further validates the model’s accuracy in assessing image quality for HD map updates. The MLLM with DPCS provides more accurate confidence scores than GCS by dynamically determining whether to account for additional parameters, ensuring a context-aware assessment. For instance, in the post-snow scene, it correctly captured snow impact (3/10) and lane visibility (7/10), yielding a confidence score of 7, closely matching human judgment, whereas GCS assigned a much lower score of 5, underestimating lane visibility. Similarly, in illumination and rainy conditions, DPCS consistently assessed visibility, producing scores closely aligned with human evaluators, while GCS often misjudged lane relevance due to its rigid scoring mechanism.

Table \ref{tab:mini} highlights the limitations of other MLLM-based methods, which assign high confidence scores based on overall image clarity rather than lane visibility. In the first image, despite high clarity, the absence of lane lines renders it unusable, yet other models score it high, proving their failure to assess lane relevance. Similarly, in adverse weather conditions, they rate images as fair or poor based on general visibility, while our MLLM with DPCS accurately evaluates lane visibility, marking its scores high, closely comparable to human assessments. These findings emphasize the superiority of our MLLM-driven Dynamic Piecewise Confidence Score Model in providing reliable, context-aware scoring for HD map updates.

\begin{table}[h]
    \centering
    \caption{Comparison of Human and Our MLLM Average Confidence Scores for Crowdsourced Vehicle Collected Data.}
    \vspace{-5.50pt}
    \label{tab:human_vlm_scores}
    \resizebox{\columnwidth}{!}{
    \def\arraystretch{1.45}
    \begin{tabular}{cccccc}
        \toprule
         {\quad}\quad & \multicolumn{3}{c}{\textbf{Average Confidence Score}} & \multicolumn{2}{c} {\textbf{Accuracy w.r.t. Human}} \\
        \cmidrule(lr){2-4} \cmidrule(lr) {5-6}
        \textbf{Total Images} & \textbf{Human} & \textbf{MLLM (GCS)} & \textbf{MLLM (DPCS)} & \textbf{MLLM (GCS)} & \textbf{MLLM (DPCS)}  \\
        \midrule
        18937 & 8.6 & 5.7 & 7.3 & 66.28\% & 84.88\% \\
        \bottomrule
    \end{tabular}
    }
\end{table}

\subsubsection{Evaluation on Real Crowdsourced Vehicle Collected Data}

Table \ref{tab:human_vlm_scores} and Figure \ref{fig:bar} demonstrate that MLLM with DPCS significantly outperforms GCS, achieving 84.88\% accuracy in aligning with human evaluators (8.6 vs. 7.3), whereas GCS lags at 66.28\% (5.7 avg. score). Unlike GCS, which underestimates lane visibility and disproportionately assigns images to the 2-5 range (6492 images), DPCS more accurately distributes scores, closely matching human evaluation, particularly in the 5-8 (7728 vs. 6628 human) and 8-10 (6886 vs. 8121 human) ranges. This dynamic, context-aware scoring approach ensures reliable HD map updates by filtering low-visibility images while retaining high-quality data, preventing misjudgments common in GCS. By selectively penalizing distortions rather than applying rigid thresholds, DPCS enhances update accuracy and ensures a more precise, human-aligned confidence estimation.

\begin{table}[h!]
\centering
\caption{Confidence Score Comparison for Images Captured in the Same Link Area but from Different Local Maps at Various Times of the Day}
\vspace{-5.50pt}
\label{tab:samelocal}
\resizebox{\columnwidth}{!}{
\begin{tabular}{llcccccc}
\toprule
{\quad}\quad & \multicolumn{4}{c}{\textbf{Images}} \\
\cmidrule{3-5}
&& \includegraphics[width=0.3\linewidth, height=0.2\linewidth]{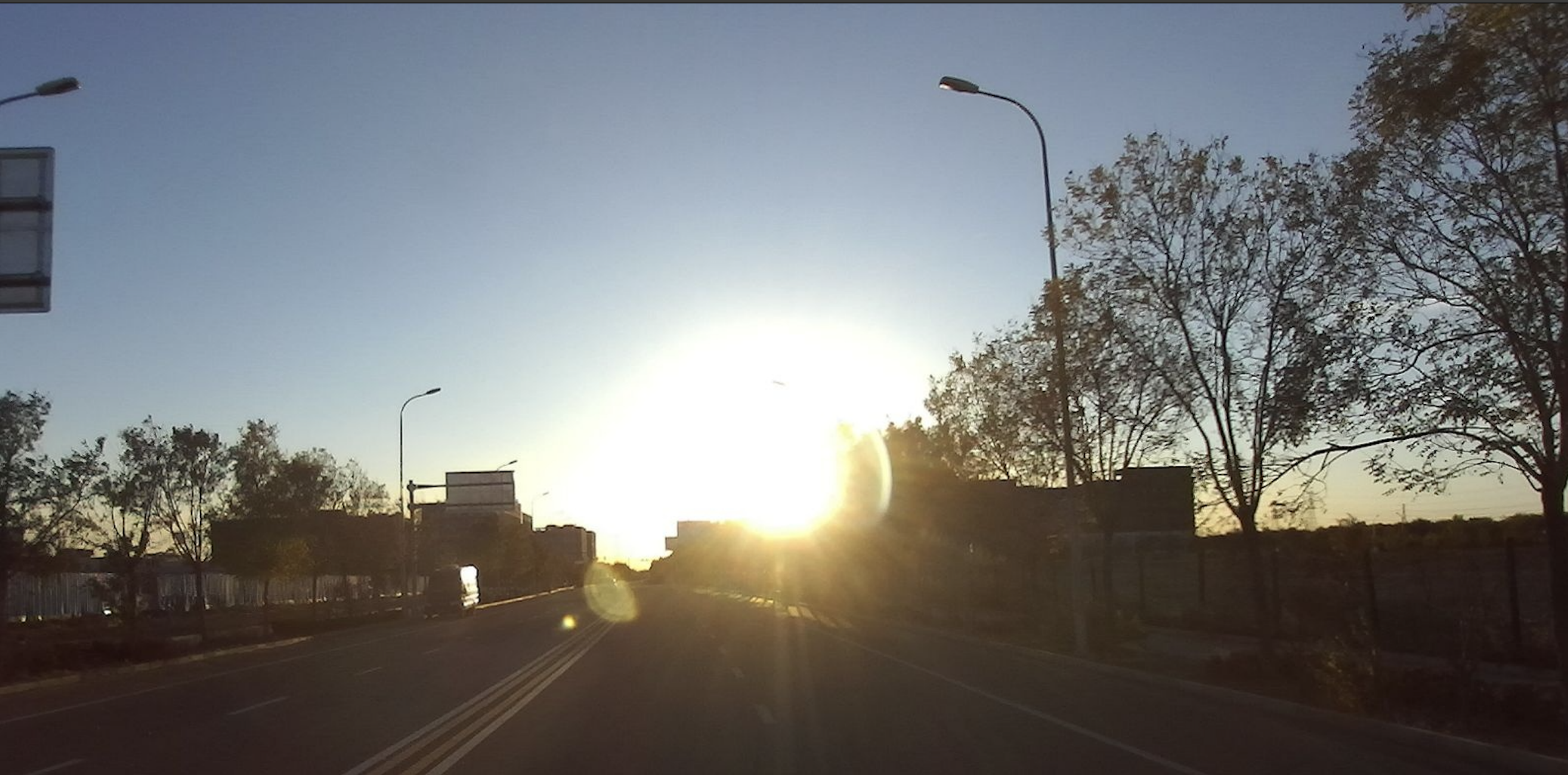} & \includegraphics[width=0.3\linewidth, height=0.2\linewidth]{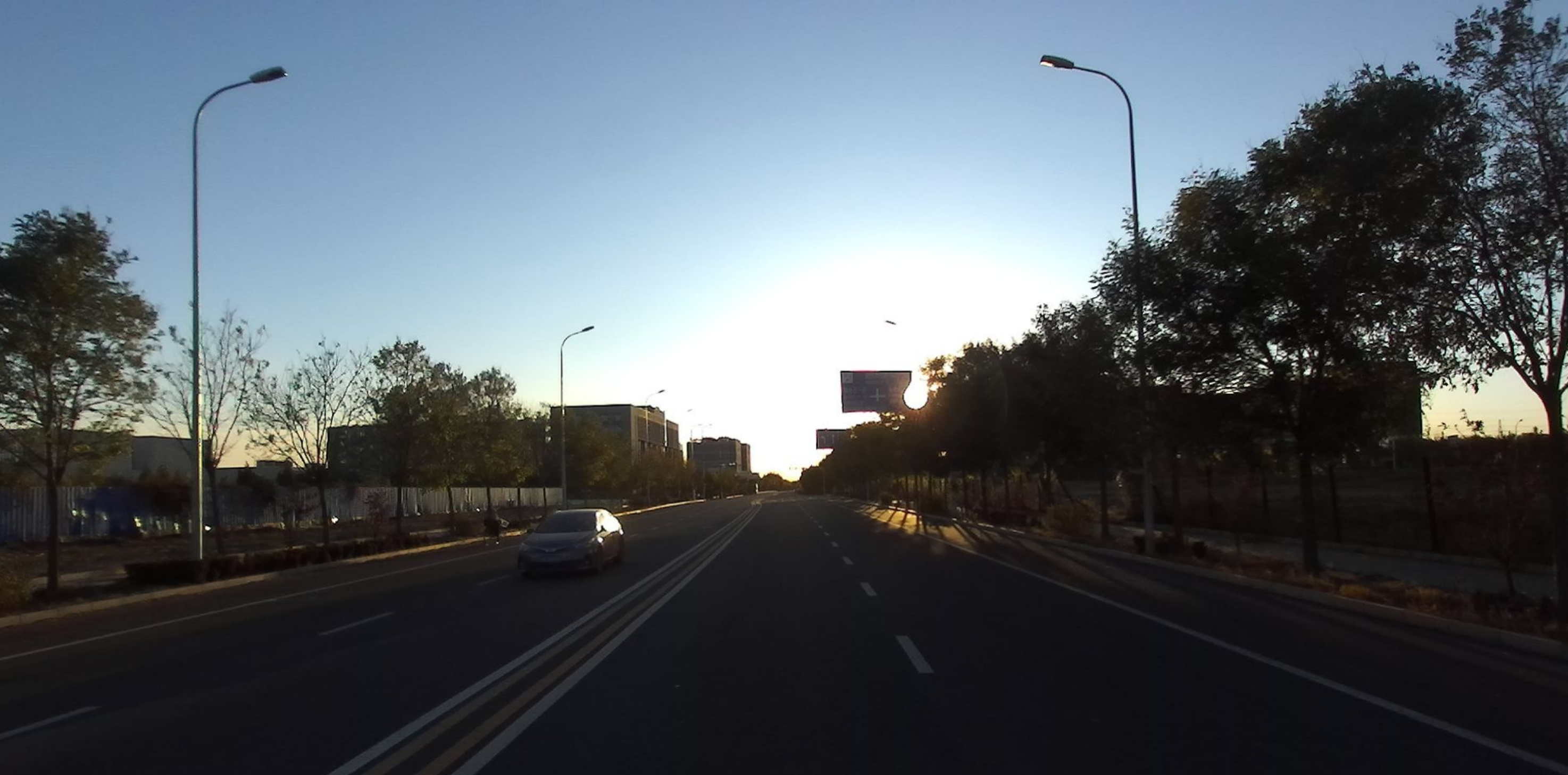} & \includegraphics[width=0.3\linewidth, height=0.2\linewidth]{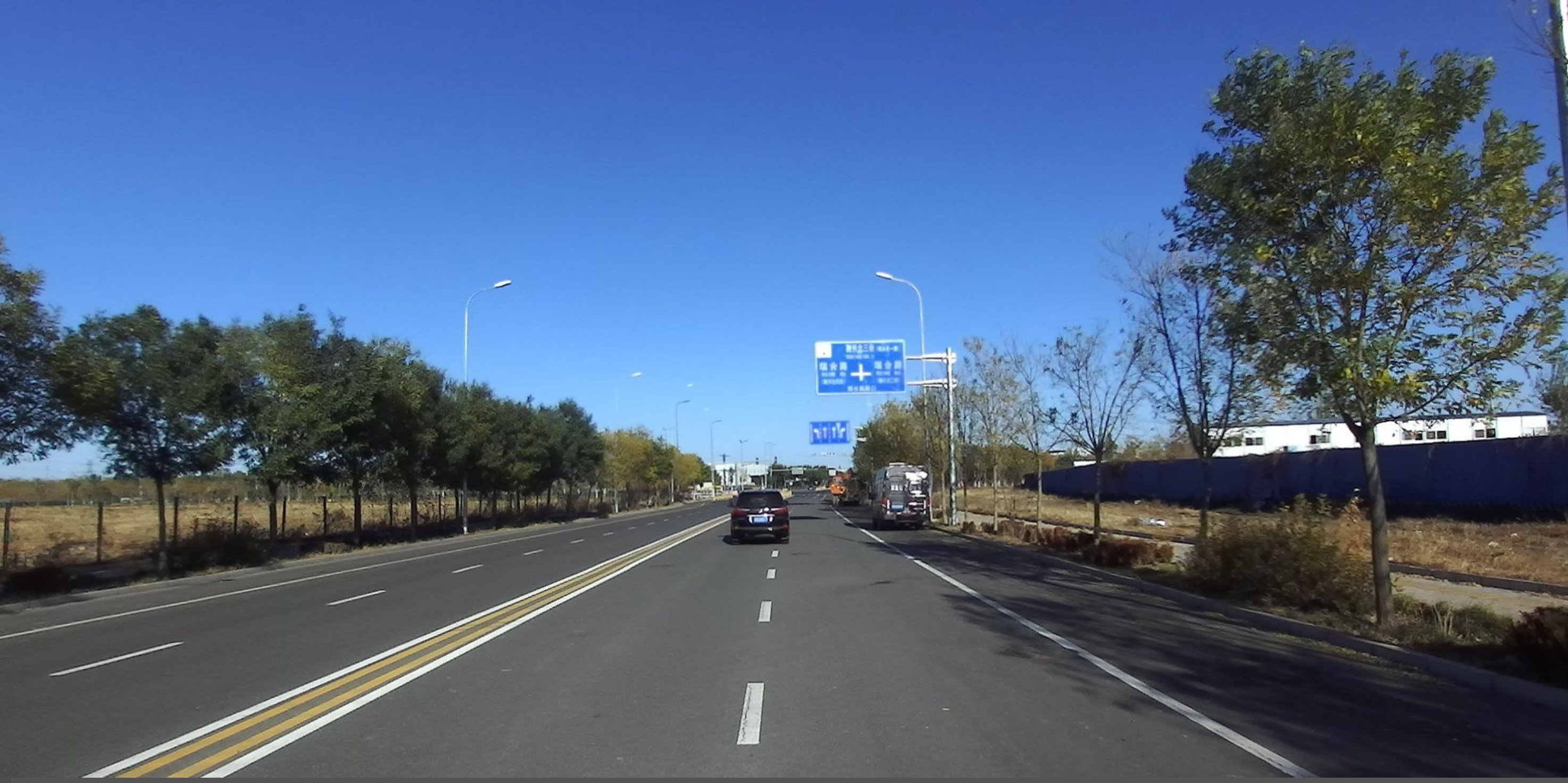} \\
\cmidrule{3-5}
{\textbf{Methods}} & \multicolumn{4}{c}{\textbf{Confidence Score}}\\ 
\midrule
MiniGPT-v2 (zero shot) \cite{minigpt} & & 3 & 4 & 9 \\
Human (GCS) & & 1.6 & 6 & 10 \\
Ours MLLM (DPCS) & & 1.8 & 6.4 & 9.8 \\
\bottomrule
\end{tabular}
}
\end{table}

\begin{table}[h!]
\centering
\caption{Average Confidence Score for Images sequences within the local maps of some specific Link Area}
\label{tab:cs}
\resizebox{\columnwidth}{!}{
\begin{tabular}{p{1.8cm}p{2.0cm}p{2.0cm}p{2.0cm}p{2.0cm}p{2.0cm}}
\toprule
{\textbf{Link Area}} & \multicolumn{5}{c}{\textbf{Average Confidence Score for Local Map (Out of 10)}} \\
\cmidrule(lr){2-6}
& \textbf{Map 1} & \textbf{Map 2} & \textbf{Map 3} & \textbf{Map 4} & \textbf{Map 5} \\
\midrule
6  & 8.80 & 8.46 & 7.82 & 6.57 & 5.38 \\
8  & 8.34 & 8.30 & 7.72 & 6.72 & 5.86 \\
21 & 7.84 & 7.49 & 7.40 & 7.19 & 6.87 \\
47 & 7.92 & 7.92 & 7.77 & 7.55 & 5.70 \\
67 & 8.35 & 7.75 & 7.25 & 5.97 & 5.96 \\
73 & 8.57 & 7.82 & 7.75 & 5.82 & 5.57 \\
\midrule
{Average} & {8.30} & {7.96} & {7.62} & {6.64} & {5.89} \\
\bottomrule
\end{tabular}
}
\end{table}

Table \ref{tab:samelocal} highlights variability in image quality across local maps within the same link area. Our MLLM closely aligns with human evaluations, ensuring precise analysis, whereas MiniGPT-v2, using a rule-based zero-shot approach, primarily assesses overall image clarity. Our model effectively penalizes overexposure (Image 1), rewards lane visibility in low-light conditions (Image 2), and provides a fine-grained assessment in optimal settings (Image 3). These findings emphasize the necessity of a confidence-driven selection process for reliable local map fusion.

Table \ref{tab:cs} validates our MLLM-driven data cleansing model, which ranks and organizes local maps within specific link areas using confidence scores. Local Maps 1–3 achieve high scores (7.62–8.30), indicating clear lane visibility, while lower scores in Local Maps 4 and 5 suggest issues like glare, blur, or adverse weather. This selective approach improves HD map accuracy and robustness for autonomous navigation.

\begin{table}[h!]
\centering
\caption{Map Update Error Comparison among Baselines vs Confidence-based vs Original MiniGPT-v2}
\label{tab:up}
\resizebox{\columnwidth}{!}{
\begin{tabular}{cccccc}
\toprule
{\textbf{Link Area}} & \textbf{Baseline} & \multicolumn{3}{c}{\textbf{Confidence-based approach}} &
\textbf{MiniGPT} \\
\cmidrule(lr){2-2} \cmidrule(lr){3-5} \cmidrule(lr){6-6}
& $e_\text{{AME}}$ (m) & Seq1 $e_\text{{AME}}$ (m) & Seq3 $e_\text{{AME}}$ (m) & Seq5 $e_\text{{AME}}$ (m) & $e_\text{{AME}}$ (m)\\
\midrule
6  & 0.39 & 0.31 & 0.30 & 0.34 & 0.33\\
8  & 0.41 & 0.29 & 0.27 & 0.36 & 0.34\\
21 & 0.40 & 0.31 & 0.29 & 0.35 & 0.36\\
47 & 0.36 & 0.32 & 0.31 & 0.34 & 0.29\\
67 & 0.33 & 0.30 & 0.28 & 0.32 & 0.30\\
73 & 0.36 & 0.26 & 0.25 & 0.31 & 0.28\\
\midrule
{Average} & {0.37} & {0.30} & {0.28} & {0.34} & {0.32}\\
\bottomrule
\end{tabular}
}
\end{table}

\begin{figure}[t]
\centering
\scriptsize
    \includegraphics[width=0.8\columnwidth]{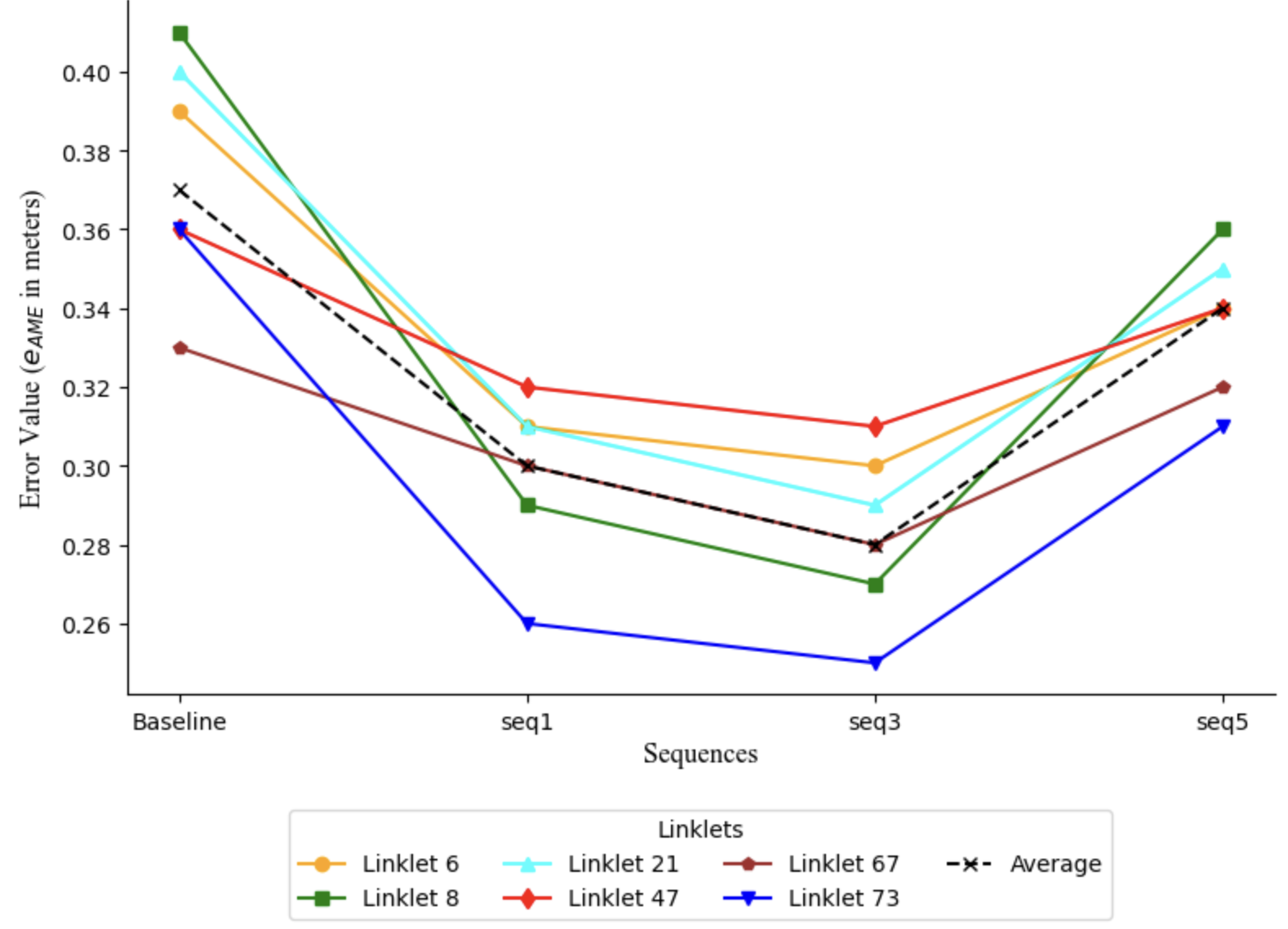}
\caption{Error Value Comparison Across Sequences for Different Link Areas.}
\label{fig:barg}
\end{figure}

\begin{table}[h!]
\centering
\caption{Average Mean Error (AME) for Lane Line Updates using Different Methods}
\label{tab:fcm}
\resizebox{\columnwidth}{!}{
\begin{tabular}{l l c}
\toprule
\textbf{Type} & \textbf{Methods} & \textbf{$e_\text{AME}$ (m)} \\
\midrule
  & Pose Graph \cite{das2020posegraph} & 0.99 \\
 Traditional Methods  & Lane Clustering \cite{kim2021hdmapupdate} & 0.65 \\
 (no filtering) & Road model + graphSLAM \cite{liebner2019crowdsourced} & 0.45 \\
 & Our Baseline (ICP \cite{besl1992icp} + DBScan \cite{ester1996dbscan}) & 0.37 \\
\midrule
 & MiniGPT-v2 \cite{minigpt} (0 shot rule-based Filtering) & 0.32 \\
 MLLM-driven  & Confidence-based approach only Top 1: Seq1 & 0.30 \\
 filtering methods & Confidence-based approach till Top 3: Seq3 & 0.28 \\
 & Confidence-based approach till Top 5: Seq5 & 0.34 \\
\bottomrule
\end{tabular}
}
\begin{tablenotes}
\scriptsize
\item Křehlík et al. \cite{Křehlík_Vanžura_Skokan_2023} stated that 0.32 is the maximum error permitted for lane lines.
\end{tablenotes}
\end{table}

Table \ref{tab:up} and Figure \ref{fig:barg} confirm that selecting top-$k$ sequences within the best confidence score minus 10\% achieves the best balance between quality and quantity for data fusion. Seq3, which fuses the top three local maps, achieves the lowest error (0.28 m), outperforming MiniGPT-v2, which filters images below a confidence score of 7 before fusion, and the baseline, which integrates all local maps without filtering. Seq1 ensures high accuracy but with fewer data points, while Seq5 introduces noise by incorporating lower-confidence maps. This confidence-driven approach enhances precision while preserving comprehensive map coverage. Table~\ref{tab:fcm} confirms that Seq3 is the most effective sequence for HD map updates, achieving the lowest mean error (0.28m), significantly outperforming the baseline (0.37m) and remaining well within the maximum permissible error of 0.32m \cite{Křehlík_Vanžura_Skokan_2023}. This confidence-driven selection of the top three local maps ensures an optimal balance between accuracy and data volume, surpassing threshold-based filtering by MiniGPT-v2 i.e. keeping only images having confidence score of 7 and above, and traditional methods that rely on unfiltered data fusion.

\section{Conclusion}
\label{sec:conclusion}

This study presents CleanMAP, a confidence-driven framework that leverages Multimodal Large Language Models (MLLMs) to enhance crowdsourced HD map updates by systematically addressing data quality inconsistencies. CleanMAP introduces a novel lane visibility-centric scoring model and a dynamic confidence-scoring function that enables precise filtering of low-quality data. Unlike conventional image quality metrics, CleanMAP prioritizes lane-specific visibility over generic clarity, significantly improving the accuracy and robustness of HD maps. Extensive real-world evaluations demonstrate that CleanMAP's confidence scores closely align with human assessments, facilitating interpretable, scalable, and automated HD map updates. The framework effectively balances data quality and quantity, outperforming both threshold-based and unfiltered aggregation approaches for HD Map Updates, thereby advancing reliable autonomous navigation.

Despite its strong performance, CleanMAP exhibits limitations under extreme lighting conditions (e.g., glare, low illumination) and partial occlusions caused by fog or snow, occasionally leading to conservative confidence estimations. Future research will focus on increasing dataset diversity and improving adaptability through refined dynamic confidence thresholds. Additionally, the integration of advanced real-time map fusion techniques and exploration of alternative encoders and LLM architectures will further enhance CleanMAP’s scalability, reliability, and performance. These developments will strengthen its deployment potential in complex, ever-changing driving environments within intelligent transportation systems.

\section{Acknowledgement}
\label{sec:acknowledgement}

This work was supported in part by the National Natural Science Foundation of China (U22A20104, 52472449), the Beijing Natural Science Foundation (L231008, L243008), and the Beijing Municipal Science and Technology Commission (Z241100003524013, Z241100003524009). This work was also sponsored by the Tsinghua University-DiDi Joint Research Center.
{
    \small
    \bibliographystyle{unsrt}
    \bibliography{main}
}

 \clearpage
\setcounter{page}{1}
\maketitlesupplementary

\section{Factors Affecting Adaptive Confidence Score Calculation}
\label{sec:keyPara}

While calculating the Adaptive Confidence Score, several factors must be considered, including key parameters affecting lane line visibility and the importance of these factors. The final confidence score of an input image is determined by dynamically assigning weights based on environmental conditions and their impact on lane visibility.

\subsection{Designing Specific Parameters to Calculate Confidence Score}

To ensure the reliability and accuracy of HD map updates, we define a set of key parameters that impact image quality and influence the visibility of lane markings. These parameters form the basis for the MLLM-based scoring system, where it assigns an individual score between 0 and 10 to each parameter based on its impact on lane visibility. The key parameters includes:

\subsubsection{Blur}
Image clarity is essential for detecting lane markings and road features. Different types of blur that affect visibility include:
\begin{itemize}
    \item Daytime Blur: Caused by motion or camera focus issues.
    \item Nighttime Blur: Often due to low light or motion in poorly illuminated areas.
    \item Streetlight-Induced Blur: Occurs when lane markings are obscured by artificial light sources at night.
\end{itemize}

\subsubsection{Illumination}
Strong lighting variations can impact image clarity, including:
\begin{itemize}
    \item Strong Sunshine or Shadows: Excessive brightness or deep shadows obscuring road features.
    \item Reflections or Glare: From reflective road surfaces or vehicles.
    \item Darkness: Low-light conditions where lane markings become less visible.
\end{itemize}

\subsubsection{Weather Conditions}
Environmental conditions can obscure road markings and reduce visibility:
\begin{itemize}
    \item Rain, Snow, and Fog: Adverse weather conditions that diminish lane visibility.
    \item Sandstorms: In desert regions, sandstorms can reduce visibility to near zero, affecting map updates.
\end{itemize}

\subsubsection{Lane Line Degradation}
Over time, lane markings may wear out and become unclear or invisible, making them unreliable for HD map updates.

\subsubsection{Obstacles Covering Lane Lines}
Vehicles, debris, or objects on the road may obstruct lane markings, making it difficult to assess road conditions accurately.

\subsubsection{Lane Line Visibility}
The overall visibility of lane markings in an image directly impacts its usability for HD map updates.

\subsection{Designing Specific Task-Guided Questions/Prompts}
To ensure the model accurately assesses image quality, a Task-Guided Instruction Prompting system is implemented. This system guides the model through structured prompts that focus on critical aspects such as lane line visibility, weather conditions, obstacles, and different types of blur. By directing the model’s attention to relevant factors, the resulting confidence scores remain contextually accurate.

\subsubsection{Task-Guided Questions for Image Evaluation}
Each prompt is designed to assess a specific aspect of the image, ensuring structured evaluation:

\begin{itemize}
    \item Question 1: Detailed Scene Description Prompt  
    "Provide a detailed description of the scene in the image, focusing on lane line visibility, the impact of vehicles or obstacles, weather conditions, and any other factors affecting clarity."

    \item Question 2: Daytime Blur Prompt  
    "Rating blurred image during daytime: [Score 0-10] - Rate the overall image clarity/sharpness on a scale of 0-10, where 10 is extremely blurry and 0 is tack sharp."

    \item Question 3: Nighttime Blur Prompt  
    "Rating blurred image during nighttime: [Score 0-10] - Rate the image clarity on a scale of 0-10, considering lane line visibility."

    \item Question 4: Streetlight Blur at Night Prompt  
    "Rating blurred lane lines due to Street Lights at Night: [Score 0-10] - Rate the clarity of lane lines, where 10 is extremely blurred and 0 is perfectly sharp."

    \item Question 5: Lane Line Invisibility due to Illumination Prompt  
    "Rating Lane Lines Invisibility due to Illumination (strong sunshine/shadows/darkness): [Score 0-10] - Rate how invisible lane lines are due to strong illumination effects."

    \item Question 6: Invisibility due to Fog Prompt  
    "Rating Lane Lines Invisibility due to Fog: [Score 0-10] - Rate the extent to which lane lines are obscured by fog."

    \item Question 7: Invisibility due to Rain Prompt  
    "Rating Lane Lines Invisibility due to Rain: [Score 0-10] - Rate how blurred lane lines are due to rain."

    \item Question 8: Invisibility due to Snow Prompt  
    "Rating Lane Lines Invisibility due to Snow: [Score 0-10] - Rate how snow obscures lane lines."

    \item Question 9: Invisibility due to Sandstorm Prompt  
    "Rating Lane Lines Invisibility due to Sandstorm: [Score 0-10] - Rate how blurred lane lines are due to sand."

    \item Question 10: Lane Line Degradation Prompt  
    "Rate the condition of lane lines on a scale of 0 to 10, where 0 is completely worn off and 10 is perfectly clear."

    \item Question 11: Vehicles Obstructing Lane Lines Prompt  
    "Rate the visibility of lane lines blocked by vehicles, where 10 is fully blocked and 0 is fully visible."

    \item Question 12: Overall Lane and Lane Marking Visibility Prompt  
    "Rate the overall visibility of the lanes and lane markings in the image on a scale of 0-10, where 10 means they are clearly visible, and 0 means they are completely invisible."
\end{itemize}

\subsection{Efficient Selection of Parameters Based on Context}
Evaluating all parameters in every scenario may lead to hallucination, where the model assigns arbitrary or inaccurate scores. To mitigate this, parameter selection is dynamically adjusted based on environmental conditions:

\begin{itemize}
    \item In clear, sunny weather, irrelevant parameters such as rain, snow, and fog are omitted to prevent unnecessary noise in the scoring process.
    \item In adverse weather conditions such as heavy rain, snow, or fog, the weights of these factors are increased due to their significant impact on image clarity. Conversely, parameters such as illumination and streetlight blur, which become less relevant, are weighted lower.
\end{itemize}

This adaptive parameter selection optimizes the model’s focus on relevant factors, reducing the risk of hallucinations and ensuring confidence scores remain accurate.

\subsection{Importance of Weight Assignment}

Each parameter affects image quality differently depending on environmental conditions. To ensure accurate confidence score calculations, parameter weights are dynamically adjusted:

\begin{itemize}
    \item In images collected during rain, snow, or fog, higher weights are assigned to weather-related parameters as these conditions obscure lane markings.
    \item In clear conditions, illumination-related factors such as reflections, shadows, and glare from the sun or streetlights are weighted more heavily.
    \item Degradation and obstacle-related parameters are assigned significant weights in all conditions, as they consistently affect lane marking detection.
\end{itemize}

The model dynamically adjusts these weights to maintain accuracy by ignoring irrelevant parameters in specific conditions. For example, fog-related parameters in clear weather are either assigned low weights or ignored to prevent unnecessary confusion in the model’s reasoning process.

Therefore, by integrating adaptive parameter weighting, task-guided instruction prompting, and context-aware parameter selection, the confidence scoring model ensures precise and reliable assessments. This approach enhances HD map update accuracy by filtering unreliable data while preserving essential information.

\section{Diverse Data Collection and Annotation for MLLM-Driven Confidence Scoring}

\subsection{Diverse Data Collection for Training}
To ensure that the model can accurately assess data quality across a wide range of conditions, a small but diverse dataset was collected, consisting of both real-world and synthetic images, as shown main paper Figure 3. This dataset includes images captured from connected and automated vehicles (CAVs) under various environmental conditions, as well as handcrafted images designed to simulate specific adverse scenarios.

The dataset is composed of two primary sources:
\begin{itemize}
    \item \textbf{Online Crowdsourced Data:} Real-world images were gathered from vehicles operating in urban and rural environments under varying weather conditions. These images include those captured in daylight and nighttime, as well as during adverse weather such as rain, snow, fog, and dust storms. This diverse set ensures that the model learns to handle a broad range of conditions that may degrade data quality.

    \item \textbf{Synthetic Images:} To supplement real-world data, synthetic images were generated to simulate extreme conditions such as heavy rain, dense fog, and severe illumination effects, including reflections, shadows, and glare from headlights and streetlights. These synthetic images enable the model to generalize to rare but critical conditions that are essential for robust HD map updates.
\end{itemize}

The dataset comprises approximately 1,000 images, including blurred images from both daytime and nighttime scenarios. This curated dataset enables the model to learn how different factors, such as blur, lighting, and weather, affect data quality. By exposing the model to a wide range of adverse conditions, the training process ensures accurate real-time data quality assessment under real-world constraints.

\begin{figure*}[h]
\centering
    \includegraphics[width=2\columnwidth]{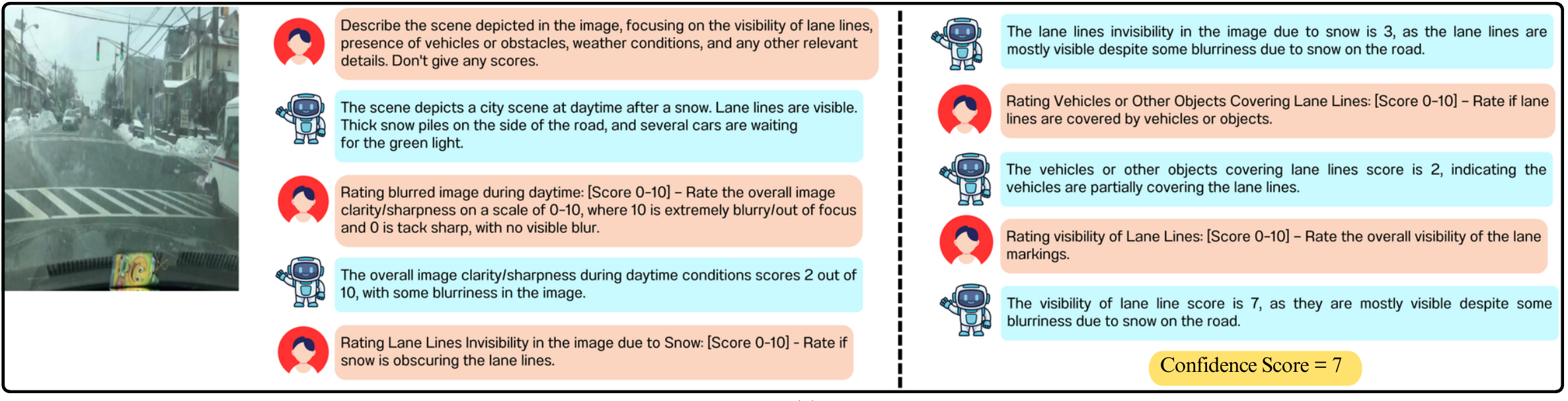}
\caption{Scenario 1: Snow Conditions with Minor Blur.}
\label{fig:s1}
\end{figure*}

\begin{figure*}[h]
\centering
    \includegraphics[width=2\columnwidth]{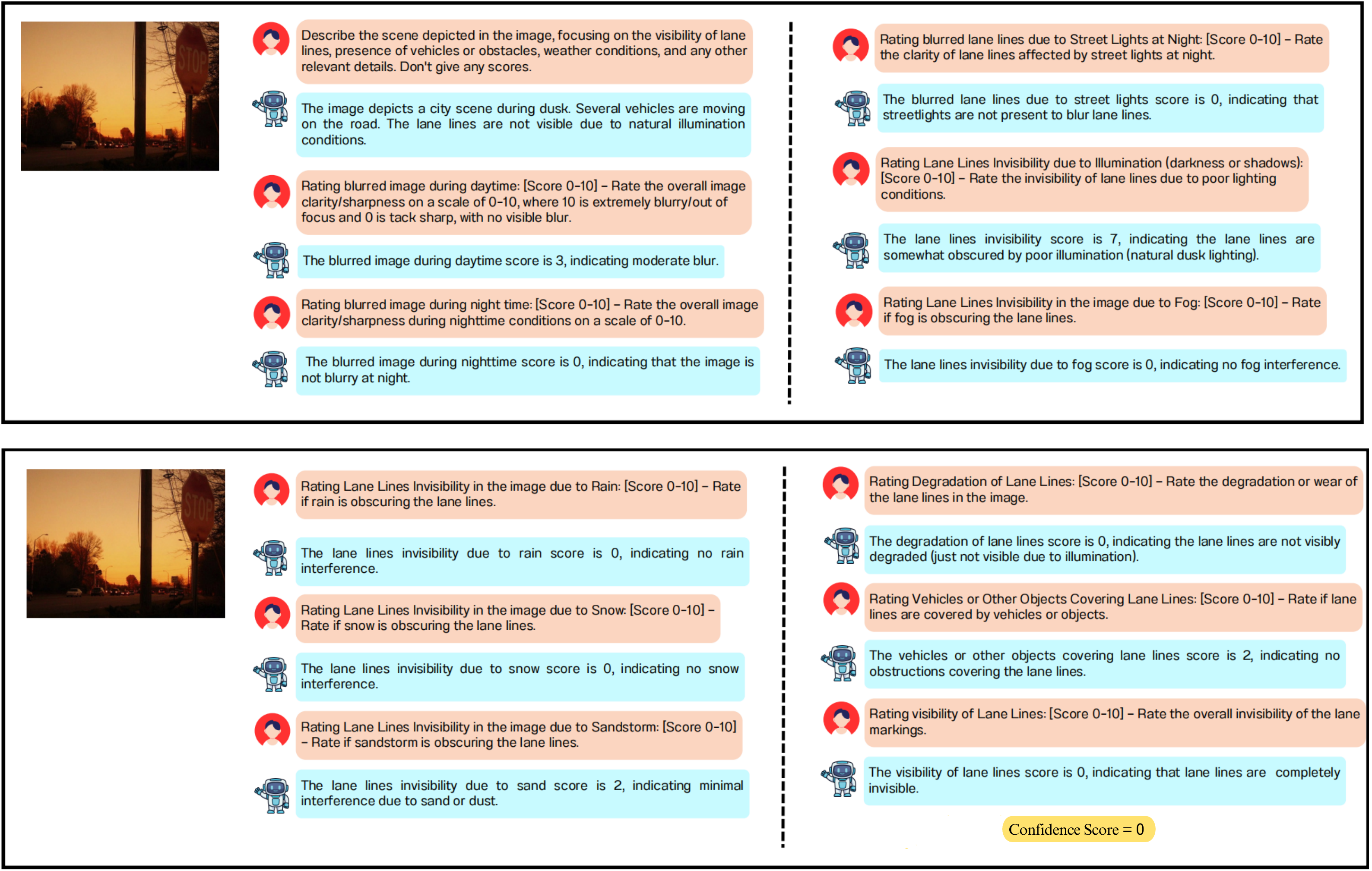}
\caption{Scenario 2: Dusk Scene with Poor Illumination. The model evaluates lane line visibility under poor lighting conditions, correctly filtering out irrelevant factors while highlighting the impact of natural illumination.}
\label{fig:s2}
\end{figure*}

\subsection{Accurate Annotation of Collected Data}
The collected data is meticulously annotated based on predefined parameters that measure image quality. Each image is manually scored across multiple factors affecting lane line visibility and overall road conditions. These annotations provide a strong baseline for the model during training.

Manual annotation enables the model to learn the relationship between visual cues and external conditions such as blur, rain, snow, and illumination, which impact road visibility to varying degrees. For instance, lane markings may become nearly invisible in dense fog but remain relatively clear in mild rain. Accurate annotations allow the model to capture these nuances by assigning well-defined scores for each factor. Without structured annotations, the model would struggle to interpret how different conditions influence lane visibility and image clarity.

For each image, multiple parameters are assessed and scored to capture how different conditions affect image quality and lane line visibility:
\begin{itemize}
    \item \textbf{Weather Conditions (Fog, Rain, Snow, Sandstorm):} Natural elements significantly impact lane line visibility. Each weather condition is scored on a scale (e.g., 0-10) to reflect its severity in the image. Precise annotation ensures the model can appropriately adjust confidence scores for images affected by these conditions.

    \item \textbf{Blur (Daytime and Nighttime):} Blur can result from camera motion, poor focus, or adverse lighting. Since its causes and impacts vary between daytime and nighttime, separate annotations for each condition are necessary to ensure proper learning.

    \item \textbf{Illumination (Sunshine, Shadows, Darkness):} Strong sunshine, deep shadows, or nighttime darkness can obscure lane markings, making them difficult to detect. Accurate annotation of illumination levels ensures the model correctly evaluates when lighting conditions affect lane visibility, adjusting the image’s usability score accordingly.

    \item \textbf{Degradation of Lane Lines:} Over time, lane markings degrade, becoming unclear or invisible. Annotating the condition of lane markings is essential, as it directly impacts the model’s ability to evaluate road geometry and lane detection reliability.

    \item \textbf{Obstacles Covering Lane Lines (Vehicles, Debris):} Vehicles, debris, or other objects that obscure lane markings should be annotated. The extent to which lane lines are blocked directly affects the accuracy of HD map updates.

    \item \textbf{Overall Lane Line Visibility:} The overall visibility of lane markings is the most critical factor in determining the usability of an image for HD map updates. This parameter encapsulates how all the aforementioned conditions collectively impact lane detection.
\end{itemize}

Thus integrating diverse real-world and synthetic data, along with precise manual annotations, the dataset ensures that the model learns to assess data quality across a wide range of challenging conditions. The structured annotation process enables the model to differentiate between various environmental factors, ultimately improving confidence scoring for HD map updates.



%
\section{Additional Qualitative Analysis of MLLM-driven Confidence Scoring}
\label{sec:supplementary}

\begin{figure*}[h]
\centering
    \includegraphics[width=2\columnwidth]{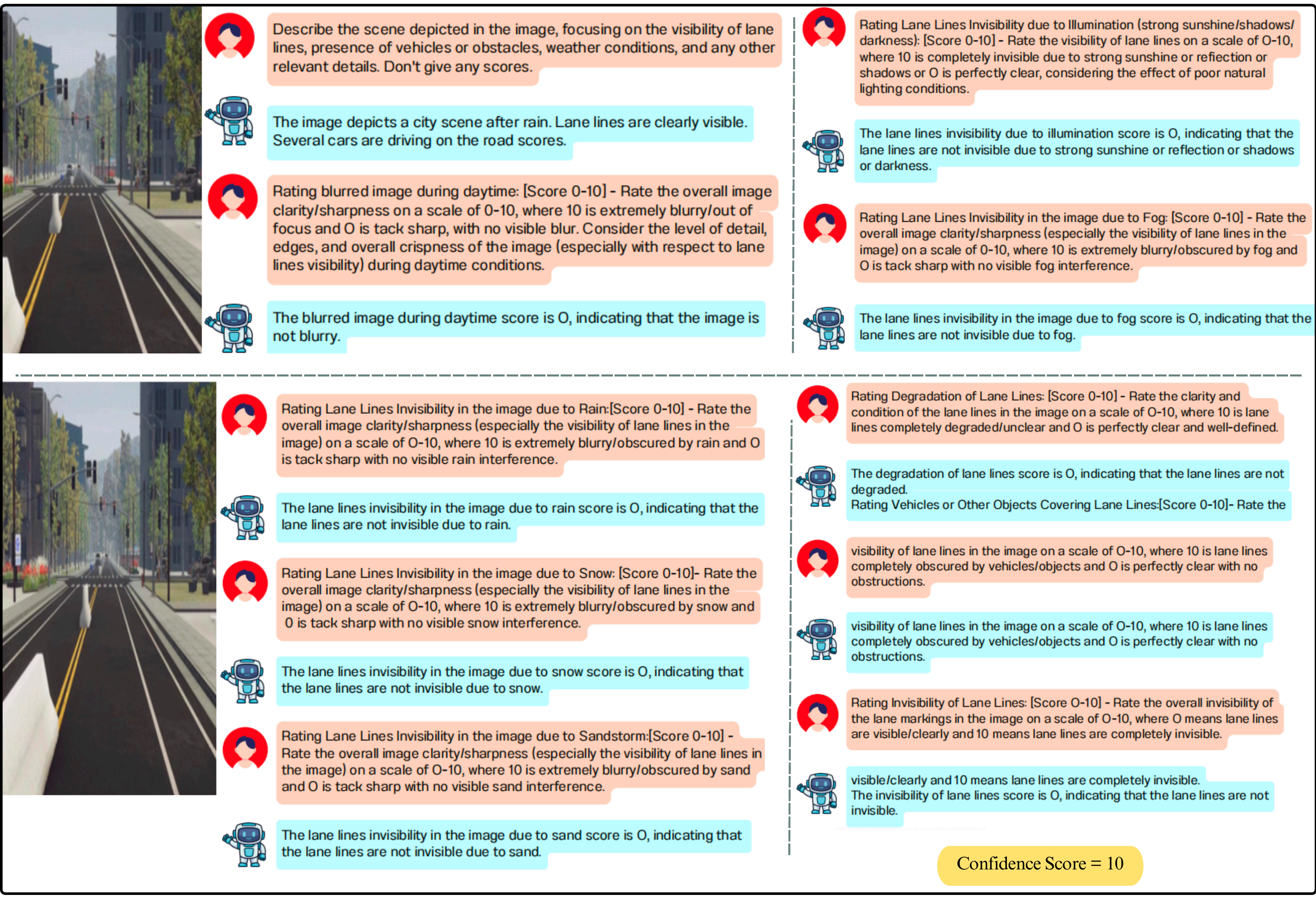}
\caption{Scenario 3: Bright Sunlight Causing Glare. The model correctly identifies intense glare and obstruction from a truck, demonstrating its efficiency in visibility scoring.}
\label{fig:s3}
\end{figure*}

\begin{figure*}[h]
\centering
    \includegraphics[width=2.0\columnwidth]{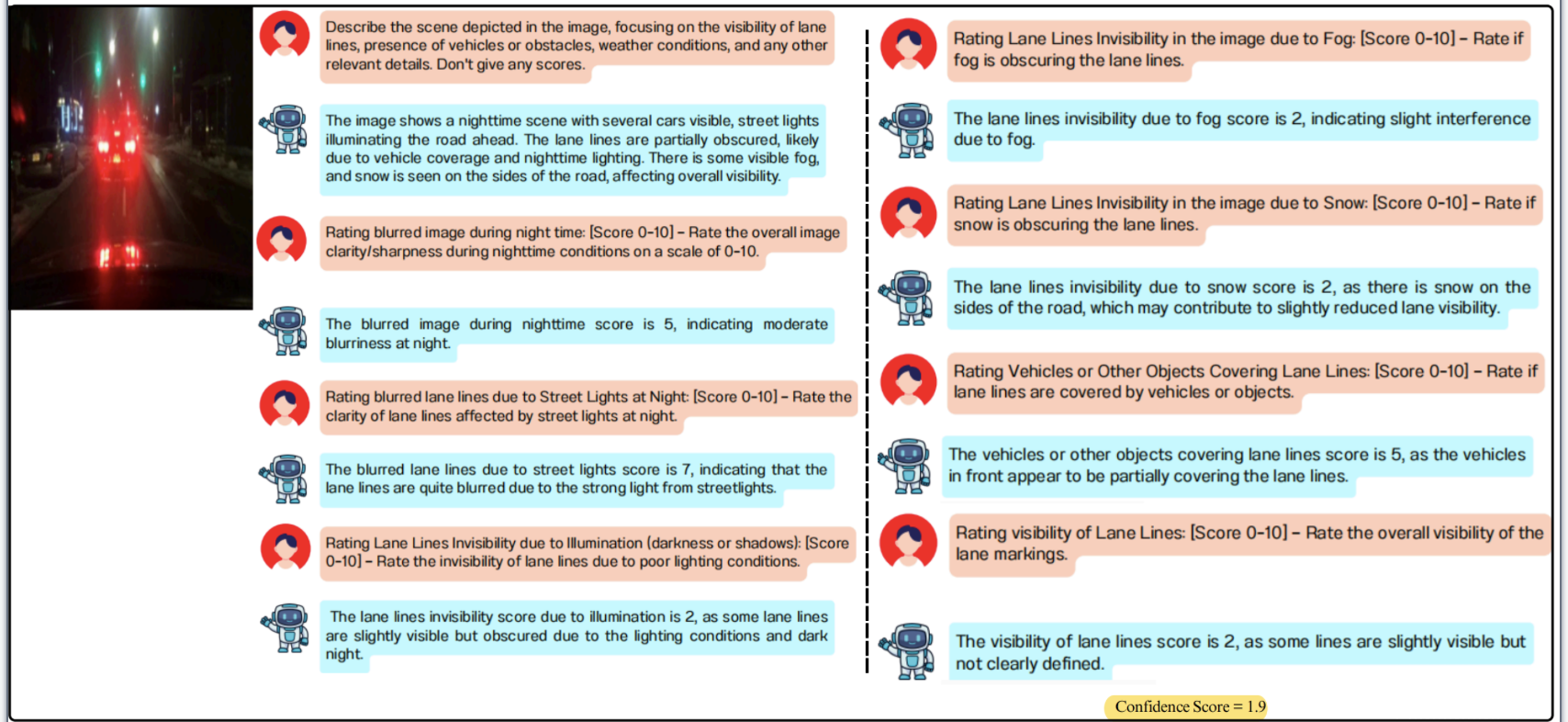}
\caption{Scenario 4: Snow-Covered Lane with Night Illumination.}
\label{fig:s4}
\end{figure*}

\begin{figure*}[h]
\centering
    \includegraphics[width=2.0\columnwidth]{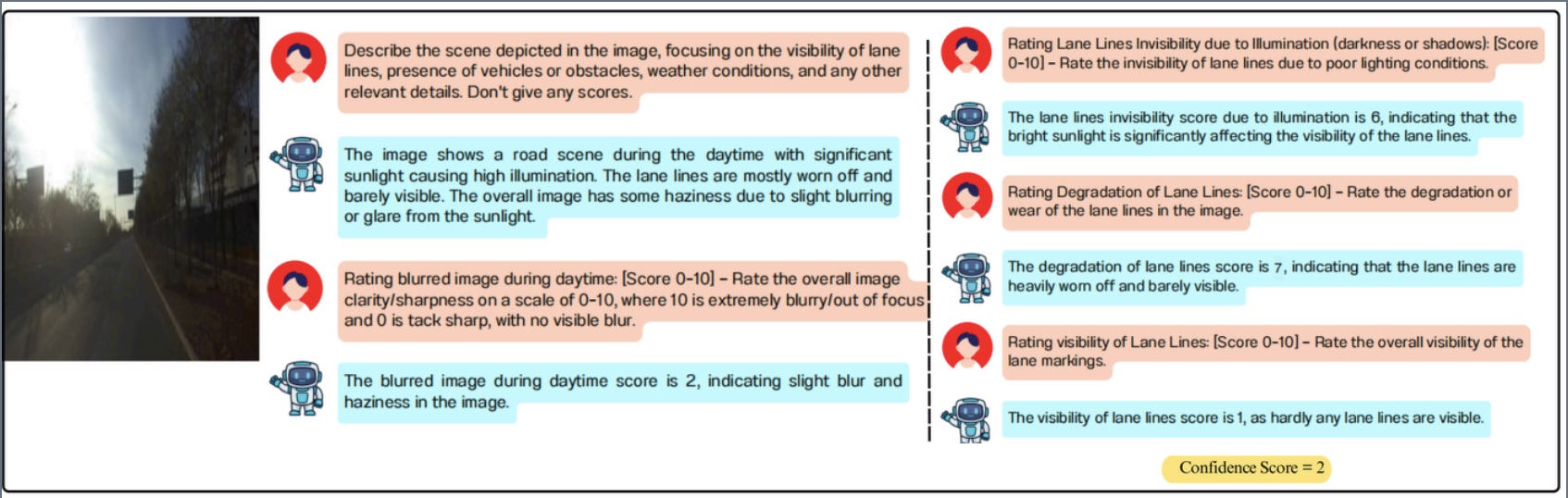}
\caption{Scenario 5: Daytime Scene with Glare and Degraded Lane Lines.}
\label{fig:s5}
\end{figure*}

\subsection{Scenario-Based Evaluations}
\label{sec:scenario-evaluations}

\subsubsection{Scenario 1: Snow Conditions with Minor Blur}
\label{sec:scenario1}
The scenario in Figure \ref{fig:s1} shows a post-snow city scene where thick snow piles are visible on the sides of the road. The lane lines are mostly visible, but there is some blur in the daytime conditions. The MLLM was able to capture the following aspects correctly:
\begin{itemize}
    \item \textbf{Blur Detection:} The model rated the image clarity as 2/10, indicating a minor blur that slightly affects lane visibility.
    
    \item \textbf{Snow Effect:} The model correctly identified snow interference but rated the lane line visibility at 3/10, indicating the lane lines are mostly visible but slightly obscured by snow. 
    
    \item \textbf{Obstruction by Vehicles:} The model recognized that some vehicles partially obstruct the lane lines ahead, giving a score of 2/10 for vehicles covering the lane lines.
    
    \item \textbf{Overall Lane Line Visibility:} The model rated lane visibility at 7/10, implying that the image provides sufficient clarity for most parts of the lane, despite the snow and slight blur.
\end{itemize}

The MLLM performed accurately in identifying the key visibility obstructions, particularly the snow and blurriness. It successfully excluded irrelevant factors like fog and nighttime blur, making this a strong example of efficient MLLM application. The confidence score of 7 reflects that the image is still usable for map updates, although the visibility could be affected by snow.

This section provides additional qualitative analysis of of the MLLM-driven confidence-scoring model, including scenario-based evaluations.

We present qualitative assessments across different environmental conditions, highlighting CleanMAP’s ability to accurately identify lane visibility challenges and compute confidence scores.

\subsubsection{Scenario 2: Dusk Scene with Poor Illumination}
\label{sec:scenario2}

In this scenario, illustrated in Figure \ref{fig:s2}, the image was captured at dusk, where poor illumination significantly affects lane line visibility. The model identified key factors impacting visibility:
\begin{itemize}
    \item \textbf{Illumination Issues:} The model rated lane line invisibility at 6/10, indicating moderate obstruction due to natural lighting conditions.
    \item \textbf{Blur Detection:} The daytime blur score was 2/10, suggesting slight distortion in the image.
    \item \textbf{Environmental Factors:} As no snow, fog, or rain were present, these parameters were correctly assigned a score of 0.
    \item \textbf{Obstruction:} No obstructions from vehicles or objects were detected, receiving a 0 score.
\end{itemize}

This scenario highlights MLLM’s capacity to accurately assess illumination-based visibility challenges while effectively filtering out irrelevant conditions. The confidence score of 1.8 confirms that the image is of low quality for map updates.

\subsubsection{Scenario 3: Bright Sunlight Causing Glare}
\label{sec:scenario3}

This scenario, depicted in Figure \ref{fig:s3}, captures a high-illumination road scene where strong sunlight causes glare and partial obstruction by a truck. The model effectively recognized:
\begin{itemize}
    \item \textbf{Illumination Problems:} The model scored lane line invisibility at 7/10, attributing poor visibility to intense glare.
    \item \textbf{Daytime Blur:} A blur score of 2/10 was assigned, indicating slight haziness due to sunlight.
    \item \textbf{Obstruction by Vehicles:} The truck partially obstructed lane markings, with a vehicle obstruction score of 4/10.
    \item \textbf{Lane Visibility:} The lane lines received a low visibility score of 3/10, confirming the compounded impact of glare and obstruction.
\end{itemize}

This scenario illustrates MLLM’s ability to distinguish between different environmental factors. The confidence score of 0.4 confirms that the image is unsuitable for HD map updates due to major glare issues.

\subsubsection{Scenario 4: Snow-Covered Lane with Night Illumination}
\label{sec:scenario4}

Figure \ref{fig:s4} presents a nighttime scenario where lane visibility is affected by snow accumulation and artificial lighting from streetlights and vehicles. The model's assessment includes:
\begin{itemize}
    \item \textbf{Nighttime Blur:} Moderate blurring due to night conditions was rated at 5/10.
    \item \textbf{Streetlight Effects:} The model assigned a high score of 7/10 to streetlight-induced visibility degradation.
    \item \textbf{Environmental Factors:} Fog and snow were correctly identified with scores of 2/10 each.
    \item \textbf{Obstruction by Vehicles:} Vehicles partially covering lane lines were scored at 5/10.
    \item \textbf{Lane Visibility:} The lane lines received a poor visibility rating of 2/10, confirming significant degradation.
\end{itemize}

This scenario demonstrates MLLM’s ability to identify multiple factors affecting visibility. The confidence score of 1.9 indicates that the image is not ideal for map updates but confirms the model's effectiveness in parameter selection.

\subsubsection{Scenario 5: Daytime Scene with Glare and Degraded Lane Lines.}
\label{sec:scenario5}

This scenario, as depicted in Figure \ref{fig:s5}, shows a road during the daytime with strong sunlight affecting the visibility of lane lines:

\begin{itemize}
    \item \textbf{Daytime Blur:} The model assigns a blur score of 2/10, indicating slight blur and haziness, which is accurate given the strong sunlight affecting the scene’s sharpness.

    \item \textbf{Illumination Problems:} Lane line invisibility due to illumination was rated at 6/10, suggesting that bright sunlight significantly affected lane line clarity.

    \item \textbf{Degradation of Lane Lines:} The model identified significant wear and tear on the lane lines, assigning a score of 7/10 for lane degradation.

    \item \textbf{Lane Visibility:} Lane lines are rated poorly for visibility, receiving a score of 1/10, as they are barely discernible due to both degradation and bright sunlight.
\end{itemize}

This scenario highlights the model’s ability to correctly assess both glare and lane line degradation. The assessment of blur, lane visibility, and the impact of sunlight is accurate, as is the degradation score. The model appropriately ignores irrelevant factors such as rain and fog, which are not present in the image. The low confidence score of 2 reflects the poor overall quality of the image for mapping purposes.

\subsection{Optimal Data Selection for HD Map Updates}
\label{sec:data-fusion}

To ensure optimal data selection for HD map updates, the confidence-driven fusion strategy prioritizes high-quality local maps. The selection process follows these principles:
\begin{itemize}
    \item \textbf{High-confidence images} are prioritized for inclusion in HD map updates.
    \item \textbf{Dynamic confidence thresholds} are used to avoid excessive filtering and ensure data sufficiency.
    \item \textbf{Environmental adaptability} ensures that the model dynamically adjusts scoring weights based on real-world conditions.
\end{itemize}

This systematic approach significantly improves the accuracy and reliability of HD maps while maintaining efficient data processing. The supplementary qualitative results confirm CleanMAP’s capability to robustly assess and score lane visibility across diverse environmental conditions. The model demonstrates strong adaptability by correctly identifying glare, poor illumination, and snow-related obstructions while filtering out irrelevant conditions. The confidence-driven scoring and data fusion approach ensures that only high-quality images contribute to HD map updates, enhancing reliability in autonomous navigation.




\section{Systematic Workflow of MLLM-Driven Confidence-Based HD Map Updates}
\label{sec:mapupdate} 

In HD map updates, integrating local map data from crowdsourced vehicles presents both opportunities and challenges. The objective is to generate a reliable global HD map by fusing individual local maps while ensuring geometric consistency, feature alignment, and positional accuracy.

Let \( M_{\text{global}} \) represent the global HD map, constructed as the union of multiple local maps \( M_{\text{local}_i} \) contributed by individual vehicles:

\begin{equation}
    M_{\text{global}} = \bigcup_{i=1}^{n} M_{\text{local}_i}
\end{equation}

where \( M_{\text{local}_i} \) consists of spatial data points \( (x_i, y_i, z_i) \), representing key road features such as lane lines and road boundaries. To fuse these maps, geometric alignment is performed to bring all local maps into a common coordinate system by minimizing deviations in overlapping data points and compensating for sensor inaccuracies and trajectory differences.

Once aligned, the final fusion step involves clustering algorithms to group closely related data points while filtering noise. This structured approach ensures an accurate, up-to-date global HD map that reflects real-time road conditions. By integrating geometric alignment and confidence-driven fusion, the model enhances HD map precision and reliability, making it highly effective for autonomous navigation.

\subsection{Optimal Local Maps Selection}

The model calculates an average confidence score for each local map, reconstructed from a sequence of timestamped images. Each image is assigned an individual confidence score by the MLLM-based Scoring model, evaluated based on environmental conditions and lane line visibility. These confidence scores are used to rank and organize local maps within specific map links, ensuring that only the most reliable data is utilized for further processing. Local maps with higher confidence scores are given preference for association. This selection process enhances the accuracy and consistency of HD map updates by prioritizing high-confidence local maps while filtering out unreliable data.

\subsection{Introducing Changes in Prior Local Maps for Future Lane Line Updates}

In real-world scenarios, local map data evolves due to road construction, lane shifts, or infrastructure modifications. To evaluate the effectiveness of the map update process, intentional modifications are introduced in prior local maps, enabling a realistic assessment of how new information is integrated into the existing HD map.

Modifications are performed through three primary tasks:
\begin{itemize}
    \item \textbf{Shifting:} Existing lane lines are shifted in the X and Y directions to simulate lane position changes due to maintenance or expansion. The original lane line is replaced by the shifted one.
    \item \textbf{Deleting:} An entire lane line is removed from the map, representing real-world lane closures or removals.
    \item \textbf{Adding:} New lane lines are introduced between existing ones to simulate road expansion or newly constructed lanes. A new lane is created by calculating the midpoint between two existing lanes with a slight offset to prevent overlap.
\end{itemize}

Once modifications are applied, the updated local map is saved and compared with the ground truth map to evaluate update accuracy. 
\begin{table}[h]
	\centering
	\caption{Definition of HD Map Element Update}
	\label{tab:update}
    \resizebox{\columnwidth}{!}{
	\begin{tabular}{ l c c c }
		\toprule
		\textbf{Update Task} & \textbf{Prior Map} & \textbf{Fused Local Map} & \textbf{Updated Map} \\
		\midrule
		Shifting & Existent & Existent & Existent \\
		Deleting & Existent & Non-existent & Non-existent \\
		Adding & Non-existent & Existent & Existent \\
		\bottomrule
	\end{tabular}
    }
\end{table}

Table~\ref{tab:update} summarizes how each modification task is performed. Before the update, tasks such as shifting and deleting apply to existing HD map elements. After the update, shifted elements retain their presence with altered positions, deleted elements are removed, and newly added elements are introduced into the HD map from the fused local map.

\subsection{Association of Modified and Reconstructed Local Map Data}

After selecting the sequences with the highest confidence scores, the model aligns them with the modified local map data using the Iterative Closest Point (ICP) algorithm. This step ensures that the reconstructed local map data points, derived from crowdsourced vehicle-collected image keyframes, are accurately aligned with the modified map data. This alignment facilitates an effective association between the two, ensuring consistency in the HD map update process.

The map association process using the Iterative Closest Point (ICP) algorithm is formulated as an optimization problem to find the optimal transformation that aligns the points in a local map \( M_{\text{local}_i} \) with those in a subsequent modified map \( M_{\text{local}_{i+1}} \). This transformation is represented as the matrix \( T \), which minimizes the alignment error between corresponding points in the two maps:

\begin{equation}
    T = \arg \min_T \sum_{i=1}^{N} \| M_{\text{local}_i} - T M_{\text{local}_{i+1}} \|^2
\end{equation}

where:
\begin{itemize}
    \item \( M_{\text{local}_i} \) and \( M_{\text{local}_{i+1}} \) represent the sets of points in the local maps before and after modification, respectively.
    \item \( T \) is the transformation matrix, consisting of a rotation matrix \( R \) and a translation vector \( t \), which aligns the two maps by minimizing positional error.
\end{itemize}

The transformation matrix \( T \) is expressed as:

\begin{equation}
    T =
    \begin{bmatrix}
        R & t \\
        0 & 1
    \end{bmatrix}
\end{equation}

where:
\begin{itemize}
    \item \( R \) is the rotation matrix.
    \item \( t \) is the translation vector.
\end{itemize}

The alignment process determines the optimal \( R \) and \( t \) by minimizing the discrepancy between corresponding points in the two maps. This ensures that changes in local map data are accurately aligned with the confidence score-based map, maintaining consistency and accuracy in the HD map update. By integrating this association process, the system incorporates the most reliable and up-to-date information into the HD map, ensuring a highly precise representation of the environment.

\subsection{Data Fusion for Map Update}

The final stage of the HD map update process involves fusing the aligned sequences with the modified local map data points. Clustering algorithms such as DBSCAN (Density-Based Spatial Clustering of Applications with Noise) are employed to fuse the data, ensuring that new and validated information is integrated into the HD map. This step accurately reflects changes in road features, lane lines, and other elements.

Map data collected from crowdsourced vehicles can vary in quality and may include noise or irrelevant information. To address this, DBSCAN is applied to cluster data points from both the associated local maps and the confidence score-based selected map. DBSCAN is particularly effective as it identifies valid clusters (e.g., lane lines and boundaries) while filtering out noise caused by sensor inaccuracies or environmental variations. The fusion of selected maps ensures that the HD map remains up-to-date and accurate.

DBSCAN is defined by two key parameters:
\begin{itemize}
    \item \textbf{Epsilon (\(\epsilon\))}: The maximum distance between two points for them to be considered part of the same cluster.
    \item \textbf{Min Samples}: The minimum number of points required to form a dense region (cluster).
\end{itemize}

Let:
\begin{itemize}
    \item \( M_{\text{local}} \) represent the local map points, where each point has coordinates \( (x_i, y_i) \).
    \item \( M_{\text{cs}} \) represent the confidence score-based selected map points with coordinates \( (x_j, y_j) \).
\end{itemize}

DBSCAN is applied to the combined dataset:
\begin{equation}
    M_{\text{combined}} = \{ M_{\text{local}}, M_{\text{cs}} \}
\end{equation}

The clustering process is formulated as:
\begin{equation}
    C = \text{DBSCAN}(M_{\text{combined}}, \epsilon, \text{min\_samples})
\end{equation}
where:
\begin{itemize}
    \item \( \epsilon \) is the neighborhood distance parameter that determines the density threshold.
    \item \( \text{min\_samples} \) is the minimum number of points required to form a cluster.
    \item \( C \) is the set of clusters generated by DBSCAN, with noise points labeled as outliers.
\end{itemize}

Therefore, by clustering valid map features and filtering out noise, DBSCAN enables the fusion of reliable map data points while discarding outliers. Unique clusters are assigned specific colors, while noise points (if any) are marked in black. This ensures that only the most accurate and up-to-date information is used in HD map updates, maintaining the integrity and precision of the map for autonomous navigation.

\section{Additional Information About Real Vehicle Crowdsourced Evaluation Data}
\label{sec:Eval} 

\subsection{Experiment Setup and Crowdsourced Data Collection}

\begin{figure}[h]
    \centering
    \scriptsize
    \includegraphics[width=0.9\linewidth]{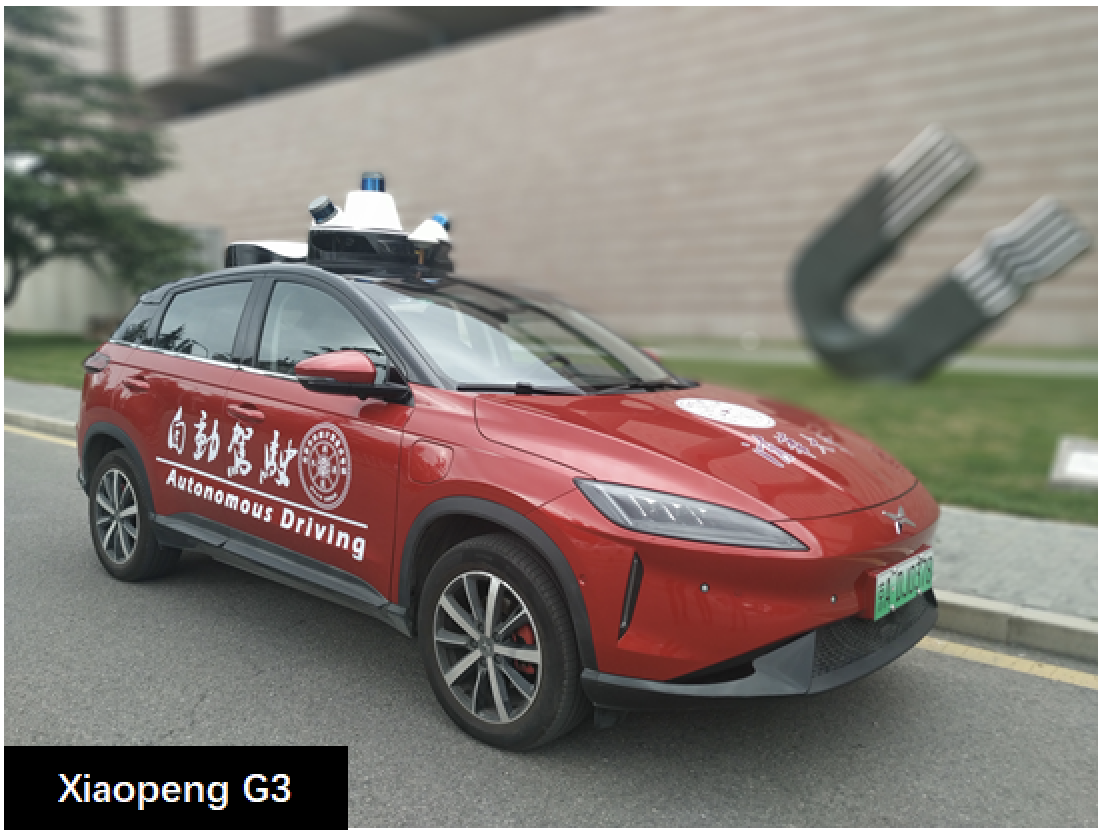}
    \caption{Real time Crowdsourced Data collection Vehicle.} 
    \label{fig:car}
\end{figure}

\begin{table}[h!]
    \centering
    \caption{Sensors equipped on the Xiaopeng G3 vehicle.}
    \resizebox{\columnwidth}{!}{
    \begin{tabular}{ccc}
    
    \toprule[1.5pt]
        \textbf{Sensor Type} & \textbf{Model} & \textbf{Parameter} \\
    \midrule[1.0pt]
        Camera & LI-AR0231-AP0200-GMSL2  & 1920$\times$1080 @28fps \\
        Lidar  & RS-Lidar-32  & 360°HFOV and 40° VFOV \\
        RTK-GNSS/IMU & NovAtel PP7D-E1 & 10cm \\
        GNSS & Ublox F9P & 10m\\
        Computer & Nuvo-6108GC & Intel i7 + Nvidia 1080 \\
    \bottomrule[1.5pt]
    \end{tabular}
    }
    \label{tab:xiaopeng}
\end{table}

This experiment utilized a real-world vehicle test platform based on the Xiaopeng G3 vehicle, as shown in Figure \ref{fig:car}. The vehicle was equipped with multiple sensors, including cameras, commercial-grade GNSS systems, and integrated inertial navigation devices. An onboard industrial-grade computer served as the central processing unit for real-time computations. A detailed list of the sensors and computing platforms used in the Xiaopeng G3 is provided in Table \ref{tab:xiaopeng}.

\begin{figure}[h]
    \centering
    \scriptsize
    \includegraphics[width=0.9\linewidth]{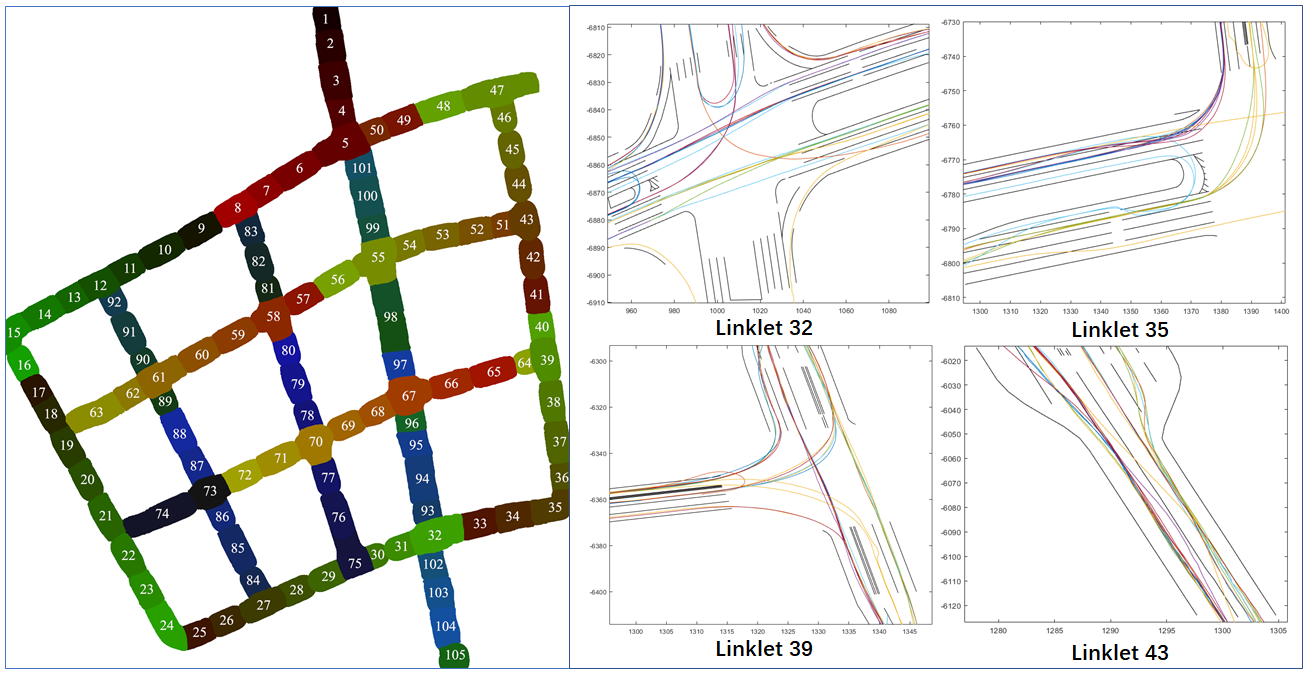}
    \caption{Global Map Representation Composed of Link Areas with Lane Line Coverage.} 
    \label{fig:map}
\end{figure}

The testing was conducted in an urban environment along roads in the Economic and Technological Development Zone, Daxing District, Beijing. The test area covered approximately 20 kilometers and included diverse urban road structures such as dual lanes, four-lane and six-lane dual carriageways, and multiple intersections with complex lane markings, including solid yellow and white lines, as well as merging and diverging lanes. Figure \ref{fig:map} illustrates the global map of the test area, which was divided into multiple Link Areas, each with specific lane line coverage. 

\begin{figure}[h]
    \centering
    \scriptsize
    \includegraphics[width=0.9\linewidth]{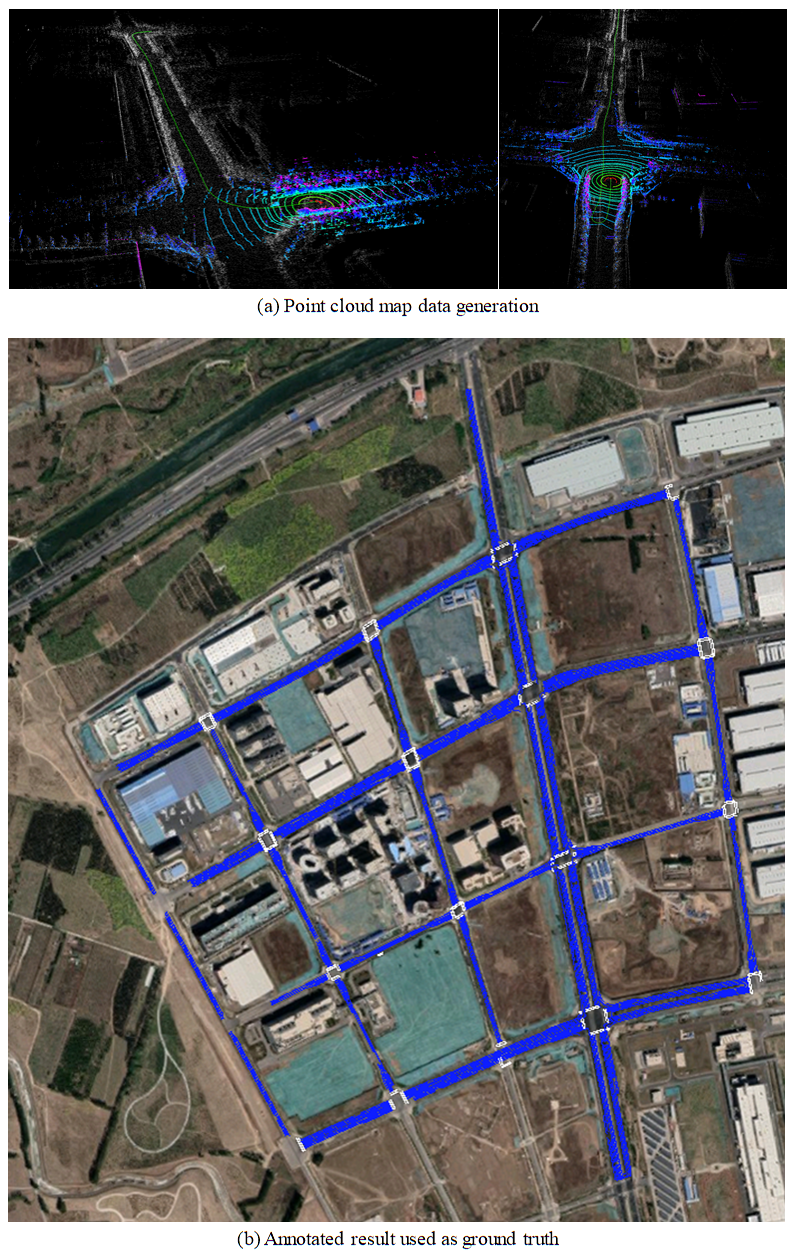}
    \caption{Generation of Groudtruth Map data.} 
    \label{fig:gt}
\end{figure}

In this area, a high-definition map was generated as the ground truth to evaluate the accuracy of the collected data. The process began with a professional survey vehicle equipped with LiDAR sensors, capturing precise point cloud data of the environment. This data formed the foundation of a high-resolution point cloud map, as illustrated in Figure \ref{fig:gt}.

\begin{table}[h!]
    \centering
    \caption{Annotated map features in the experimental data.}
    \resizebox{\columnwidth}{!}{
    \begin{tabular}{ccc}
    \toprule[1.5pt]
        Map Elements & Total Number & Annotation Format \\
    \midrule[1.0pt]
        Road Lines & 1536 & 3D Line \\
    \bottomrule[1.5pt]
    \end{tabular}
    \label{tab:GT_abc}
    }
\end{table}
\paragraph{}

Subsequently, essential map features relevant to autonomous driving, such as lane markings, were manually annotated onto the point cloud map. The annotated map provided ground truth data containing road surface elements, ensuring a reliable reference for comparison. Table \ref{tab:GT_abc} outlines the specific map features, including the total number of road lines represented as 3D line annotations.

To simulate real-world conditions, data collection was conducted using a crowdsourced approach, leveraging observations from multiple vehicles at various times and angles. Crowdsourced data provides a diverse range of perspectives, capturing dynamic changes in road conditions. The data collection process spanned eleven months, from October 2022 to September 2023, with data gathered daily between 9 AM and 5 PM to ensure broad temporal diversity. Randomized routes were selected to maximize coverage across the mapped area, allowing the system to capture multiple perspectives of the same location under different conditions.

This approach ensures a comprehensive dataset that accurately reflects variations in road structures, lane visibility, and environmental conditions, contributing to a more reliable and up-to-date HD map.

\subsection{Significance of Confidence Score}

The confidence score is crucial for determining which data can be reliably integrated into the map update process. Scores range from 0 to 10, with higher scores indicating better data quality. The classification of confidence scores is as follows:

\begin{itemize}
    \item \textbf{Confidence Score \(\sim\)2:} These images are of very low quality, often affected by extreme lighting, severe weather conditions such as rain or fog, or obstructions. Such images contribute little to accurate mapping and are generally deemed unreliable.
    
    \item \textbf{Confidence Score \(\sim\)5:} Images in this range are of moderate quality. While they may capture some useful information, they often contain partial obstructions, uneven lighting, or slightly blurred lane markings, limiting their reliability for precise mapping.
    
    \item \textbf{Confidence Score \(\sim\)9:} Images in this range are considered high quality. Captured under ideal environmental conditions, they provide clear visibility of lane markings, traffic signs, and other essential map elements, making them highly reliable for HD map updates.
\end{itemize}

\begin{table}[h!]
\centering
\caption{Visual Explanation of what each Score signifies in terms of Image Quality}
\label{tab:table_image}
\setlength{\tabcolsep}{10pt} 
\renewcommand{\arraystretch}{1.5} 
\resizebox{\columnwidth}{!}{
\begin{tabular}{llccc}
\toprule
 & \multicolumn{4}{c}{\textbf{Images}} \\
\cmidrule{3-5}
& & \includegraphics[width=0.25\linewidth, height=0.2\linewidth]{5.png} & \includegraphics[width=0.25\linewidth, height=0.2\linewidth]{6.png} & \includegraphics[width=0.25\linewidth, height=0.2\linewidth]{7.png} \\
\textbf{Parameters} & & \textbf{MLLM} & \textbf{MLLM} & \textbf{MLLM} \\ 
\midrule
Blur (Daytime)    & & 1 & 0 & 0 \\
Illumination      & & 5 & 2 & 1 \\
Degradation       & & 0 & 0 & 0 \\
Object            & & 0 & 1 & 1 \\
Visibility        & & 3 & 6 & 10 \\
\midrule
\textbf{Confidence Score} & & 1.8 & 6 & 9.6 \\
\bottomrule
\end{tabular}
}
\end{table}

The table \ref{tab:table_image} illustrates examples of images corresponding to confidence scores of 1.8, 6, and 9.6, providing a visual explanation of what each score signifies in terms of image quality.

\subsection{Key Parameters Selection}

Since the crowdsourced data is collected during the daytime under clear weather conditions, i.e., without rain, snow, fog, or sand, the primary quality check parameters used in the evaluation are:

\begin{itemize}
    \item Blur (Daytime)
    \item Illumination
    \item Lane Line Degradation
    \item Presence of Objects on the Lanes
    \item Visibility of Lane Lines
\end{itemize}

\section{Detailed Discussion on Results of Real Crowdsourced Vehicle Collected Data}
\label{sec:detail}

\subsection{Analysis of Confidence Scores for Local Maps}

The MLLM-driven confidence scoring model plays a crucial role in filtering out low-quality image sequences, directly impacting the accuracy of HD maps. Table \ref{tab:cs} provides insights into the quality of image sequences captured across different link areas in multiple local maps, as determined by the average confidence score.

\begin{itemize}
    \item \textbf{High Confidence Scores for Local Maps 1–3:}  
    The first three local maps in each link area consistently exhibit high confidence scores, ranging between an average of 7.62 and 8.30. These results suggest that images in these maps were captured under favorable conditions, where key parameters such as lane line visibility, illumination, and the absence of blur were optimal. This indicates clear lane markings and suitable weather and road conditions, making this data highly reliable for HD map updates.
    
    \begin{itemize}
        \item In Link Area 6, Local Map 1 has a high score of 8.80, Local Map 2 scores 8.46, and Local Map 3 scores 7.82, indicating optimal data quality.
        \item Similarly, Link Area 73 shows a strong confidence score of 8.57 for Local Map 1, followed by 7.82 and 7.75 for Local Maps 2 and 3, confirming good lane visibility and image clarity.
    \end{itemize}

    \item \textbf{Confidence Scores in Local Maps 4 and 5:}  
    A decline in confidence scores for Local Maps 4 and 5 across most link areas suggests challenges in the data collection process. The scores for these maps range between 5.38 and 6.87, indicating image quality degradation, likely due to high illumination, glare, or lane degradation.

    \begin{itemize}
        \item In Link Area 6, the confidence score drops from 7.82 in Local Map 3 to 5.38 in Local Map 5. This reduction could be attributed to excessive glare, blurred imagery, or obscured lane lines, making the data less reliable for HD map updates.
        \item Similarly, Link Area 67 experiences a decline from 7.25 in Local Map 3 to 5.96 in Local Map 5, suggesting deteriorating conditions such as blurred images, poor lane visibility, or traffic obstructions.
    \end{itemize}

    \item \textbf{Filtering Data Using Confidence Scores:}  
    The confidence scores generated by the Data Cleansing Model help identify local maps with valid and high-quality image sequences suitable for HD map updates. By selecting an appropriate threshold, unreliable maps can be filtered out. For instance, setting a threshold confidence score of 7.0 ensures that any local map below this value is excluded from the HD map update process.

    This approach demonstrates that:
    \begin{itemize}
        \item Only the highest-quality data is used for map updates, increasing the overall accuracy and reliability of the HD map.
        \item Maps with confidence scores above the threshold are captured under favorable conditions, ensuring good lane visibility, minimal blur, and optimal illumination.
    \end{itemize}
\end{itemize}

\subsection{Integration of Detected Changes into the HD Map}

\begin{figure}[h]
    \centering
    \scriptsize
    \includegraphics[width=1.0\linewidth]{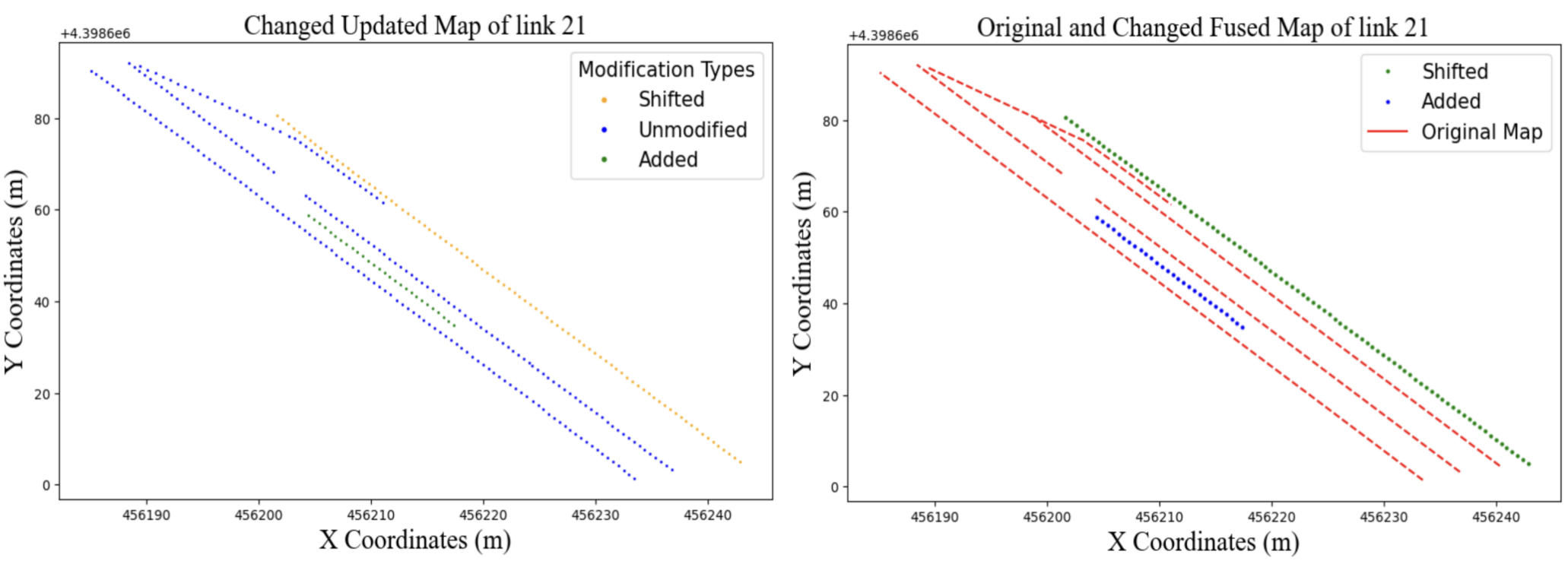}
    \caption{Visualisation of Changed New Map and Fused Map in Link Area 21.} 
    \label{fig:map1}
\end{figure}

\begin{figure}[h]
    \centering
    \scriptsize
    \includegraphics[width=1.0\linewidth]{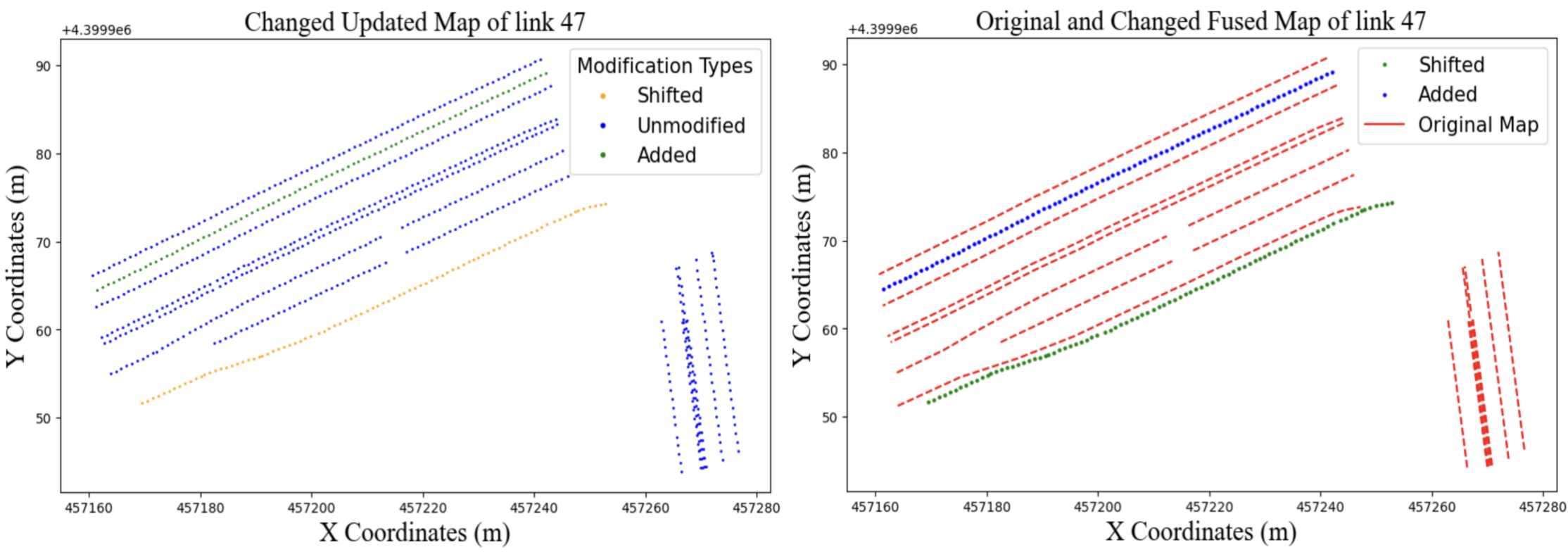}
    \caption{Visualisation of Changed New Map and Fused Map in Link Area 47.} 
    \label{fig:map2}
\end{figure}

\begin{figure}[h]
    \centering
    \scriptsize
    \includegraphics[width=1.0\linewidth]{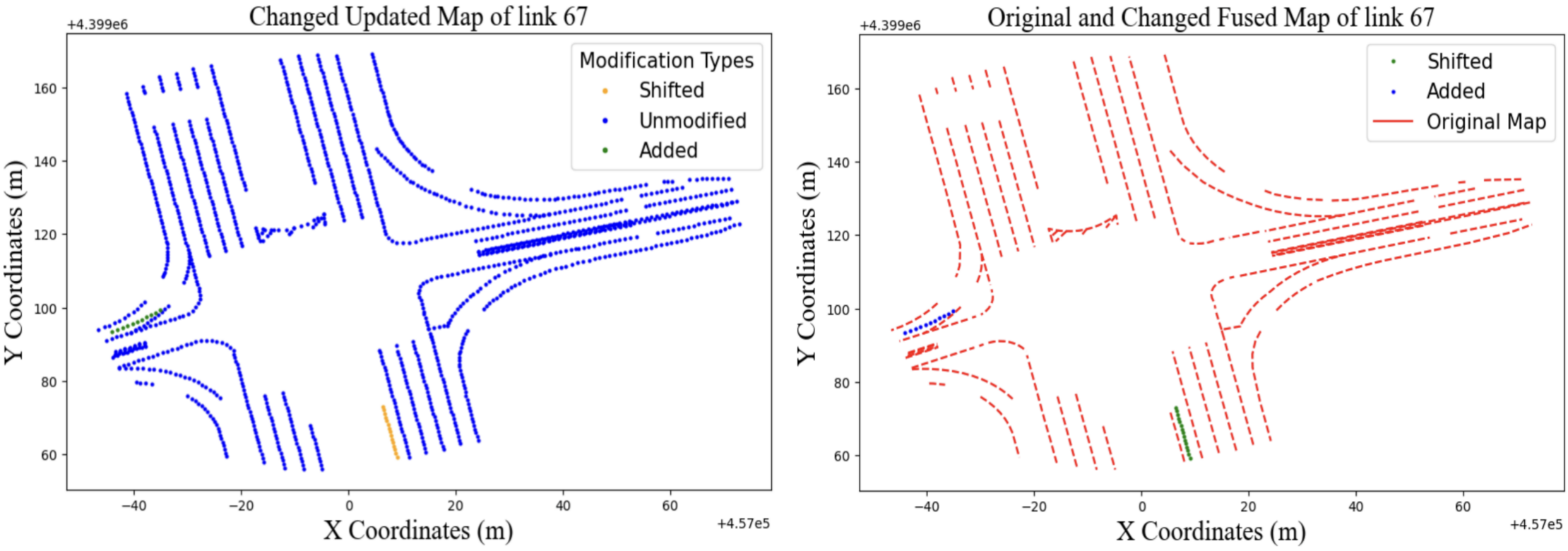}
    \caption{Visualisation of Changed New Map and Fused Map in Link Area 67.} 
    \label{fig:map3}
\end{figure}

The next phase of the HD map update process focuses on incorporating detected changes into the existing map for several link areas, as illustrated in Figures \ref{fig:map1}, \ref{fig:map2}, and \ref{fig:map3}. Three link areas were selected where specific modifications were identified and then fused with local maps containing high-confidence image sequences to generate an updated map. These changes primarily involve shifting lane lines, which includes the removal of outdated lane markings and the addition of new ones. 

The ICP-based association ensures that shifted lanes are accurately aligned with the original map, while DBSCAN handles fusion by incorporating new lanes and removing obsolete ones. Such updates are crucial for maintaining accurate and safe navigation, particularly in dynamic environments where road conditions frequently change.

In Figures \ref{fig:map1}, \ref{fig:map2}, and \ref{fig:map3}, the left side represents the Changed/Modified New Map, while the right side displays the Fused Map, highlighting the differences. The yellow color in the modified map signifies lane shifts, while green indicates newly added lanes. In the fused map, green represents lane shifts, and blue represents newly added lanes. The selected link areas, each exhibiting unique structural characteristics and modifications, emphasize the importance of these updates.

The fused map integrates the high-confidence local map with detected changes, ensuring an accurate representation of the current road layout.

\begin{itemize}
    \item \textbf{Link Area 21:} As shown in Figure \ref{fig:map1}, Link Area 21 exhibits a pattern of shifted and newly added lanes. The detected changes (left) indicate that several lane lines in the upper-left quadrant required shifting, while new lanes were added toward the center. The fused updated map (right) demonstrates the successful integration of these modifications. Since Link Area 21 includes high-traffic areas, confidence score-based data selection played a crucial role in selecting optimal sequences, minimizing discrepancies with the ground truth map, and ensuring precise lane alignment.

    \item \textbf{Link Area 47:} As shown in Figure \ref{fig:map2}, Link Area 47 demonstrates more extensive modifications. The detected changes (left) indicate multiple shifted lanes, particularly in the bottom-right region. The fused map (right) illustrates how these updates were integrated, with new lane lines reflecting the current road layout. Accurate local map sequences were essential in ensuring that the modifications aligned correctly with the updated lane configurations.

    \item \textbf{Link Area 67:} As shown in Figure \ref{fig:map3}, Link Area 67 has a complex structure with multiple intersections and branching lane lines. The detected changes (left) reveal multiple shifted lane lines, particularly near intersections, which required precise alignment. The fused map (right) integrates these changes smoothly, accurately reflecting the new configurations and enhancing routing information for navigation. In such complex environments, confidence score-based map selection is vital to capturing intricate details like lane transitions and merges with high precision.
\end{itemize}

\subsection{Physical Meaning of the terms Seq1, Seq3 and Seq5}
\label{sec:meaning}
In the evaluation presented in Table \ref{tab:up}, a sequence Seqk refers to the fusion of top k local maps ranked based on their average confidence scores while performing map update:

\begin{itemize}
    \item \textbf{Baseline:} Updates are performed by fusing all available local maps, irrespective of their confidence scores. This approach includes both high and low-quality data, maximizing the number of data points but potentially reducing overall map reliability.
    
    \item \textbf{Seq1:} Updates are made using only the local map with the highest confidence score, ensuring that only the most reliable data is used. This sequence involves the minimum number of data points but maintains the highest data quality.
    
    \item \textbf{Seq3:} The map is updated by fusing top 3 local maps i.e. first using the local map with the highest confidence score, then incorporating fusion of the second- and third-best ranked maps respectively. This sequence increases the number of data points while maintaining relatively high data quality.
    
    \item \textbf{Seq5:} The map is updated by fusing top 5 local maps i.e., from top 1 till top 5 highest scoring local maps. While this increases the data points further, the inclusion of lower-confidence maps introduces lower-quality data into the update process.
    
    \item \textbf{MiniGPT:} Updates are performed by retaining all images and their corresponding lane line data points with confidence scores of 7 or higher. This ensures that only data from local maps containing images above the confidence threshold (considered to represent good quality) is used. The confidence scores are calculated by MiniGPT-v2, which follows predefined rules emphasizing image clarity.
\end{itemize}

\subsection{Optimal Sequence for HD Map Update}

Based on the evaluation results in Table \ref{tab:up}, \ref{tab:fcm} and Figure \ref{fig:barg}, Seq3 is identified as the most optimal sequence for updating the HD map. It balances data quality and quantity, ensuring that the updated map remains accurate while incorporating sufficient data points to handle complex road configurations. This sequence offers several advantages:

\begin{itemize}
    \item \textbf{Higher Data Quality:} Seq3 maintains a high average confidence score of 7.6, which is close to the best possible score of 8.3.
    
    \item \textbf{Sufficient Data Points:} By incorporating three local maps, Seq3 ensures that enough data points are included to accurately model the map without introducing excessive noise.
    
    \item \textbf{Low Error:} The average mean error across all link areas for Seq3 is 0.28 meters, significantly lower than the baseline as well as other methods' average mean error of 0.37 meters, demonstrating that the system maintains accuracy even with increased data points.  Furthermore, the error for Seq3 falls well below the minimum accuracy requirement for lane lines in HD maps, which is less than equal to 0.32 meters, as established by Křehlík et al. (2023) \cite{Křehlík_Vanžura_Skokan_2023}
\end{itemize}

\subsection{Trade-off Between Confidence Score and Data Points}

One of the key insights from the evaluation results is the trade-off between confidence scores and the number of data points. Confidence scores indicate the quality of an image sequence, where higher scores correspond to optimal conditions such as clear lane visibility, minimal blur, and proper illumination. However, increasing the number of data points often requires incorporating local maps with lower confidence scores, introducing noise and reducing overall accuracy.

This trade-off is evident in the performance of different sequences:

\begin{itemize}
    \item \textbf{Seq1:} Utilizes only the local map with the highest confidence score, ensuring minimal error (0.30 meters on average). Although it limits the number of data points, the high data quality results in accurate map updates.
    
    \item \textbf{Seq3:} Expands on Seq1 by incorporating the second and third best-scoring local maps. This increases the number of data points while maintaining a low error rate of 0.28 meters, making Seq3 the most optimal configuration. It effectively balances data quality with sufficient data points for accurate map updates.
    
    \item \textbf{Seq5:} Adds lower-confidence local maps (fourth and fifth), leading to an increase in error to 0.34 meters. While it introduces more data points, the inclusion of lower-quality maps degrades overall accuracy.
\end{itemize}

The trade-off demonstrates that while adding more data points can improve map coverage, incorporating lower-confidence maps introduces errors. Managing this balance is crucial for maintaining both accuracy and coverage in HD map updates.

\subsection{Optimal Confidence Score Threshold for HD Map Updates}

Another critical observation from the results is the importance of setting an appropriate confidence score threshold for selecting data in HD map updates. As shown in Table \ref{tab:cs}, the first three local maps in each link area have high confidence scores ranging from 7.6 to 8.8, correlating with lower error rates in Seq1 and Seq3. In contrast, local maps 4 and 5, with confidence scores between 5.8 and 6.6, introduce greater errors in Seq5.

Based on these findings, it is recommended to set a confidence score threshold of 7.0 or higher for HD map updates. This threshold ensures that only high-quality data is used, reducing the likelihood of introducing errors due to lower-quality data. A threshold of 7.0 effectively balances data quality with the number of data points, as evidenced by Seq3, which achieves an optimal configuration.

\subsection{Reliability of the MLLM-Driven Confidence Score-Based HD Map Update}

The results demonstrate that the confidence score-based HD map update system is highly reliable. The system consistently outperforms the baseline in all cases, with Seq1 and Seq3 achieving significantly lower errors. Even Seq5, despite incorporating lower-confidence data, performs better than the baseline, proving that the confidence score-based approach effectively filters poor-quality data while utilizing high-quality inputs.

The robustness of this approach lies in its ability to minimize noise by prioritizing high-confidence local maps, ensuring highly accurate updated maps. The error values obtained from Seq1 and Seq3 meet the minimum accuracy requirements for HD maps, further affirming the reliability of this approach for real-world applications in autonomous vehicle navigation.

The confidence score-based HD map update system effectively maintains the accuracy and reliability of HD maps. The trade-off between confidence score and the number of data points is a crucial factor in the update process, and the results indicate that Seq3 provides the best balance between these two elements. By using the top three local maps with the highest confidence scores, Seq3 ensures both high data quality and sufficient data points, leading to accurate map updates.

Furthermore, the analysis supports setting a confidence score threshold of 7.0 to ensure that only high-quality data contributes to map updates. This threshold minimizes the introduction of errors while maintaining comprehensive map coverage. The results confirm that Seq3 provides the optimal configuration for HD map updates, achieving a mean error significantly lower than the baseline and meeting the accuracy requirements for autonomous navigation systems.

The confidence score-based approach not only enhances accuracy but also ensures the reliability of HD maps, making it an ideal solution for large-scale HD map updates. This system strengthens the ability of autonomous vehicles to navigate complex road environments with precision and safety.

\section{Supplementary Conclusion}
\label{sec:chapter_conclusion}

The evaluation of the confidence score-based HD map update system has provided several key insights into optimizing map accuracy using quality-assessed data. The experiments demonstrated that the proposed approach consistently outperforms the baseline, showing a significant reduction in mean errors across different sequences. Notably, Seq3 was identified as the most optimal configuration, achieving a mean error of 0.28 meters, compared to the baseline of 0.37 meters. By using the top three local maps based on their confidence scores, Seq3 managed to strike a balance between data quality and quantity, ensuring comprehensive coverage while maintaining high precision.

These findings confirm the efficacy of confidence score-driven methodology for large-scale HD map update systems, supporting safer and more precise autonomous vehicle navigation. This framework sets a strong foundation for improving autonomous vehicle navigation through more accurate and adaptive map updates.

\end{document}